%% file: saucd.tex
\documentclass[10pt,twocolumn,letterpaper]{article}

\usepackage{cvpr}              

\input{preamble}

\definecolor{cvprblue}{rgb}{0.21,0.49,0.74}
\usepackage[pagebackref,breaklinks,colorlinks,citecolor=cvprblue]{hyperref}

\def\paperID{****} 
\def\confName{CVPR}
\def\confYear{2024}

\title{Spectrum AUC Difference (SAUCD): Human-aligned 3D Shape Evaluation}

\newcommand{\authorskip}{\hspace{12mm}}
\author{
 Tianyu Luan$^{1,2}$\thanks{Work done while the first author was an intern at OPPO US Research Center.} \authorskip Zhong Li$^{2}$ \authorskip Lele Chen$^{2}$ \authorskip
 Xuan Gong$^{3,2,1}$ \\ Lichang Chen$^{2,4}$ \authorskip Yi Xu$^{2}$ \authorskip Junsong Yuan$^{1}$\\[3mm]
 {\small $^1$State University of New York at Buffalo ~~~~~~ $^2$OPPO US Research Center} \\
  {\small $^3$ Harvard Medical School~~~~~~ $^4$University of Maryland, College Park} \\
 {\tt\small \{tianyulu,jsyuan,xuangong\}@buffalo.edu} ~~~ {\tt\small\{zhong.li,lele.chen,yi.xu\}@oppo.com} ~~~  {\tt\small bobchen@umd.edu} \\
 {\small \url{https://bit.ly/saucd}}
}

\usepackage{times}
\usepackage{epsfig}
\usepackage{graphicx}
\usepackage{amsmath}
\usepackage{amssymb}
\usepackage{enumitem}

\usepackage{booktabs}
\usepackage{color}
\usepackage{diagbox}
\usepackage{microtype}
\usepackage{makecell}

\usepackage{colortbl}
\usepackage{url}
\usepackage{appendix}

\def\model{SAUCD}
\def\fname{Spectrum AUC Difference}
\def\ffname{Spectrum Area Under the Curve Difference}
\def\dsname{\textit{Shape Grading}}

\usepackage[capitalize]{cleveref}
\crefname{section}{Sec.}{Secs.}
\Crefname{section}{Section}{Sections}
\Crefname{table}{Table}{Tables}
\crefname{table}{Tab.}{Tabs.}

\makeatletter

\newcommand{\Rmnum}[1]{\textcolor{red}{\expandafter\@slowromancap\romannumeral #1@}}
\newcommand{\supl}[1]{Appendix {#1}}
\newcommand{\meq}[1]{main paper Equation (\textcolor{red}{#1})}
\newcommand{\msec}[1]{main paper Section \textcolor{red}{#1}}

\newcommand{\mtab}[1]{main paper Table \textcolor{red}{#1}}
\makeatother

\begin{document}
\maketitle
\begin{abstract}

Existing 3D mesh shape evaluation metrics mainly focus on the overall shape but are usually less sensitive to local details. This makes them inconsistent with human evaluation, as human perception cares about both overall and detailed shape. In this paper, we propose an analytic metric named \ffname{} (\model{}) that demonstrates better consistency with human evaluation. To compare the difference between two shapes, we first transform the 3D mesh to the spectrum domain using the discrete Laplace-Beltrami operator and Fourier transform. Then, we calculate the Area Under the Curve (AUC) difference between the two spectrums, so that each frequency band that captures either the overall or detailed shape is equitably considered. Taking human sensitivity across frequency bands into account, we further extend our metric by learning suitable weights for each frequency band which better aligns with human perception. To measure the performance of \model{}, we build a 3D mesh evaluation dataset called \dsname{}, along with manual annotations from more than 800 subjects. By measuring the correlation between our metric and human evaluation, we demonstrate that \model{} is well aligned with human evaluation, and outperforms previous 3D mesh metrics. 
\end{abstract}


\section{Introduction}
\label{sec:intro}

With the recent progress of 3D reconstruction and processing techniques, 3D mesh shapes have increasing applications in fields such as video games, industrial design, 3D printing, etc. In these applications, assessing the visual quality of the 3D mesh shape is a crucial task.
To meet the requirements of various applications, a promising evaluation metric should not only reflect the geometry measurement but also align with human visual perception.
Considering that human beings perceive 3D meshes in both overall shape and local details, it is a challenging task to find an evaluation metric that can align well with humans.

\begin{figure}[t]
  \centering
   \includegraphics[width=1\linewidth]{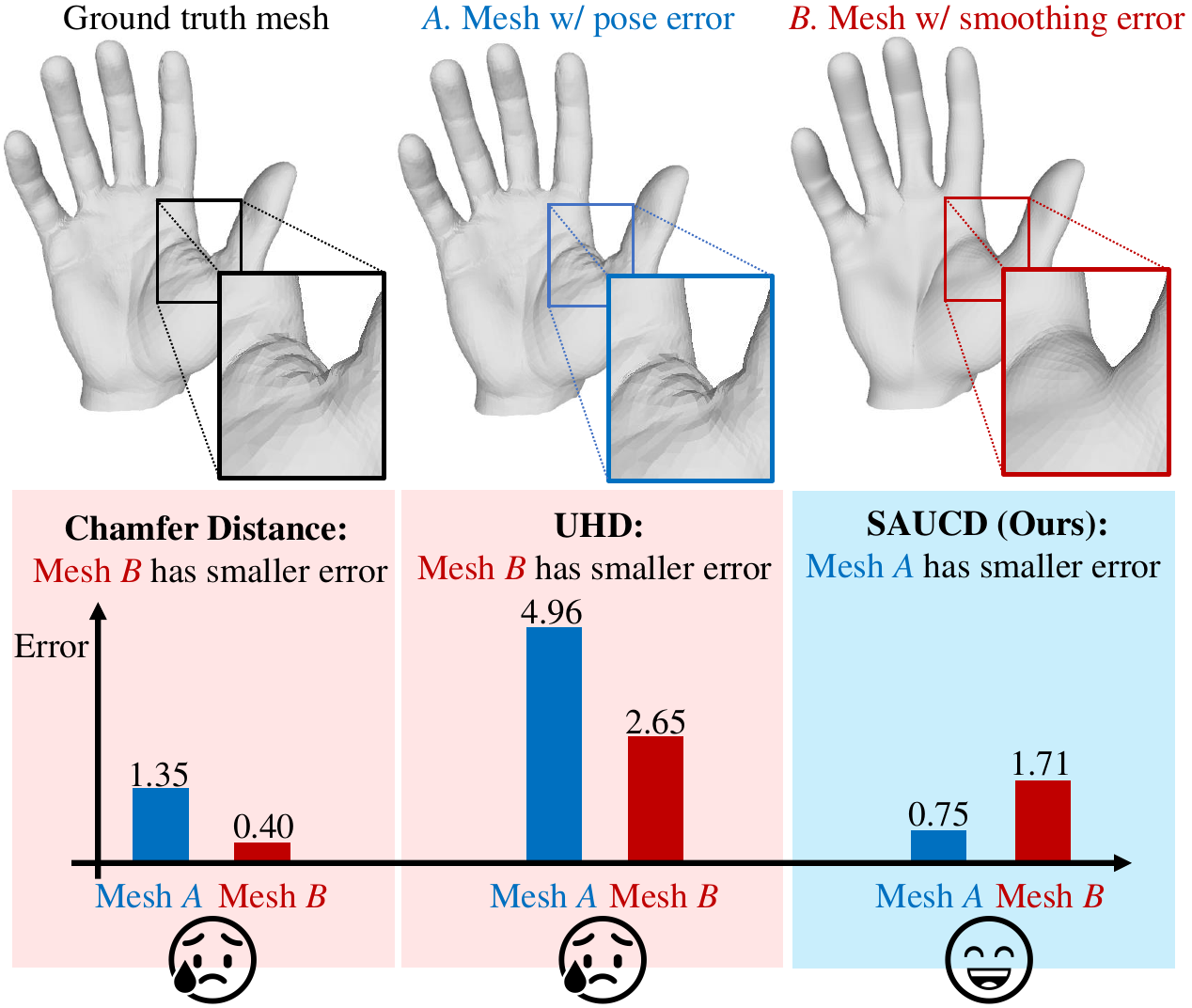}
   \caption{An example of how previous spatial domain 3D shape metrics (Chamfer Distance~\cite{borgefors1984CD} and UHD~\cite{wu2020uhd}) deviate from human evaluation. we create \textcolor{blue}{Mesh $A$} by adding a small pose error to the ground truth mesh, and by applying a large smoothing kernel to ground truth, we create  \textcolor{red}{Mesh $B$}. Contrary to human perception, previous spatial domain metrics evaluate \textcolor{red}{Mesh $B$} better than \textcolor{blue}{Mesh $A$}.
   This indicates that while they are sensitive to general shape differences, they tend to overlook high-frequency details. Note that different metrics use different units of measurement. 
   }
   \label{fig:intro}
\end{figure}

Previous metrics have the following disadvantages in this scenario.
Traditional spatial domain measurements such as Chamfer Distance~\cite{borgefors1984CD} which calculates the mean distance between a vertex on one mesh and its nearest vertex on the other mesh, can accurately measure the spatial distance. However, it does not guarantee capturing all shape details. In fact, such measurements in the spatial domain often overlook finer shape details, as the details tend to get overwhelmed by the overall shape.
\cref{fig:intro} illustrates the discrepancy between spatial measurements and human evaluation as mesh details change. Specifically, When we remove the wrinkles from the ground truth mesh (resulting in \textcolor{red}{Mesh \emph{B}}), the errors detected by previous metrics are not as significant as when we slightly change the pose of the hand (\textcolor{blue}{Mesh \emph{A}}). However, humans tend to sense a significant difference between ground truth and \textcolor{red}{Mesh \emph{B}}, but barely recognize the difference between ground truth and \textcolor{blue}{Mesh \emph{A}}. To mitigate this problem, previous works propose learning-based approaches, such as Single Shape Fréchet Inception Distance (SSFID)~\cite{wu2022ssfid} based on learnable features from 3D shape. They compare the difference between the test mesh and the ground truth mesh in the latent feature space, and the design is expected to better align human perception. 
However, such learning-based methods would require a large amount of data to train the network. Their accuracy and generalizability are limited by the size of the dataset, data distribution, and annotation quality, not to mention the potential bias in collecting human perception feedback, which could mislead the learned metrics. An analytic metric that can better explain the shape difference is thus preferred.

To address the above limitations, we design an analytic-based 3D shape evaluation metric named \ffname{} (\model{}).
Our metric measures mesh shape differences with a balanced consideration of both overall and detailed shape, making it better aligned with human evaluation. 
To allow our metric to capture detail variations, we leverage the 3D shape spectrum to decompose different levels of shape details from the overall shape, with details corresponding to higher-frequency components. The advantage of transforming the shape signal into the spectrum domain is that the high-frequency details are explicitly separated from the low-frequency overall shape. Therefore, it provides appropriate consideration to the information in different frequency bands, not just the low-frequency information of the overall shape in the dominant place. Thus, the details that human perception cares about will be better represented. Besides, the frequency analysis method allows the metric to be mostly analytic and better explained.

We design \model{} following the above inspiration. To begin with, both the test mesh and the ground truth mesh are transformed from the spatial to the spectrum domain using the discrete Laplace-Beltrami operator (DLBO), which encodes the mesh geometry information into a semidefinite Laplacian matrix. Once in the spectrum domain, we compare the regions under the two spectrums. Our \ffname{} metric is defined as the area of the non-overlapping region under the two spectrums -- a larger area indicates a greater difference.
Moreover, to better align with human evaluation, we further extend our design by learning a spectrum weight for \model{}. However, different from previous learning-based approaches that use deep networks, large datasets, and extensive learning processes, our learning-based method requires the training of a weight vector. This vector measures the sensitivity of human perception across frequency bands, making the learned metric better aligned with human perception.
We then evaluated the effectiveness of \model{} on our provided user study benchmark dataset named \dsname{}. Using \dsname{}, we compare our metrics with previous metrics by calculating the correlation between each metric and human scoring.
In summary, our contributions are listed as follows.

\begin{itemize}[noitemsep,topsep=0pt]
    \item We design an analytic-based 3D mesh shape metric named \fname{} (\model{}), which evaluates the difference between a 3D mesh and its ground truth mesh. Our metric considers both the overall shape and intricate details, to align more closely with human perception.
    \item We further extend our design to a learnable metric. The extended metric explores the human perception sensitivity in different frequency bands, which further improves this metric.
    \item We build a user study benchmark dataset named \dsname{} which is annotated by more than 800 subjects. The provided dataset verifies that both versions of our metrics are consistent with human evaluation and outperform previous methods. This dataset can also facilitate 3D mesh metric evaluation in future research.
\end{itemize}

Our experiments show that both \model{} and its extended version outperform previous methods with good generalizability to different types of objects.

\section{Related Works}
\label{sec:related}

\textbf{Metrics in 3D mesh reconstruction.} 
Chamfer Distance~\cite{borgefors1984CD} is a popular metric used in 3D mesh reconstruction tasks such as those in~\cite{lin2022cd, wei2021cd, zuo2021cd, zeng2022cd, shrestha2021cd, rakotosaona2021cd, zhang2023cd, kulikajevas2022cd}. Other spatial domain metrics, such as 3D Intersection over Union (IoU) in~\cite{hu2021iou, chen2021iou, nie2020totaliou, henderson2018iou, tang2022iou, santhanam2023iou}. F-score in~\cite{wang2018fscore, genova2020fscore, bechtold2021fscore, tatarchenko2019fscore}, and Unidirectional Hausdorff distance (UHD) in \cite{wu2020uhd} are commonly focused on the geometry accuracy of mesh shapes. These metrics can provide accurate geometry measurements, but they are not designed to align with human evaluation.
Deep-learning-based methods such as Single Shape Fréchet Inception Distance \cite{wu2022ssfid} are also used in 3D reconstruction. While these metrics have the capacity to adapt from human evaluation, they are more like black boxes, with performances subject to dataset size and annotation bias. Moreover, most previous works miss out on user study validation to verify if their metrics align with human evaluation.

\begin{figure*}[t]
  \centering
   \includegraphics[width=1\linewidth]{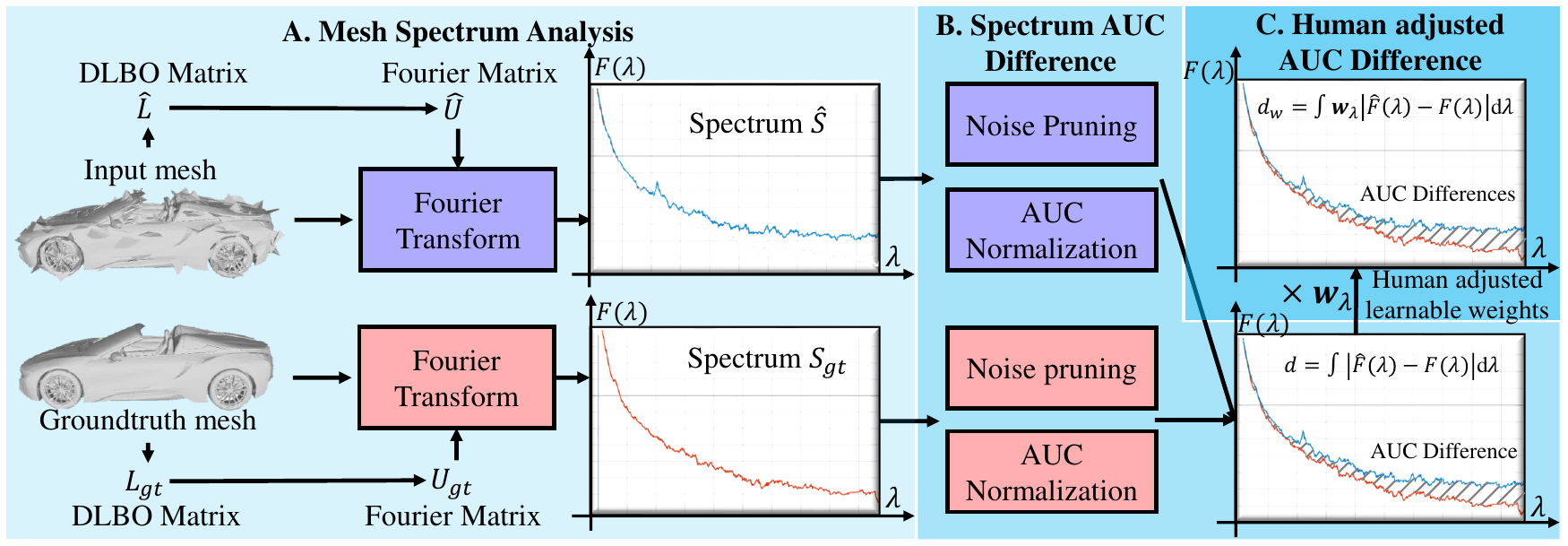}
   \caption{Our \model{} metric is designed as follows: \emph{A.} We use mesh Fourier Transform to analyze the spectrums of test and ground truth mesh. 
   \emph{B.} We compare the difference between two spectrum curves by calculating the Area Under the Curve (AUC) difference.
   \emph{C.} We further extend our metric by multiplying the AUC difference with a learnable weight to capture human sensitivity in each frequency band.}
   \label{fig:pipeline}
\end{figure*}

\textbf{3D shape generation metrics.}  Multiple metrics have been used in 3D shape generation, such as  Minimal Matching Distance (MMD)~\cite{achlioptas2018mmd}, Jensen-Shannon Divergence (JSD)~\cite{kullback1951jsd}, Total Mutual Difference (TMD)~\cite{wu2020uhd}, Fréchet Pointcloud Distance (FPD) \cite{shu2019fpd}, \etc{}.
These metrics are designed to measure the differences between the generated distributions, while our task is to build a metric to compare the shape of two meshes.

\textbf{3D mesh compression and watermarking metrics.} Previous works~\cite{wang2010wm,bulbul2011wm,lavoue2009wm,corsini2013wm} focused on evaluating mesh errors in mesh compression and watermarking. Since compression and watermarking pursue mesh errors that cannot be detected by humans, they mainly focus on barely noticed errors. However, our task is to build a metric that can handle generally occurring errors that happen in 3D reconstruction tasks and applications.

\section{Proposed Method}
\label{sec:method}
Our task is to design a metric aligned with human evaluation to measure the shape difference between a test triangle mesh and its corresponding ground truth triangle mesh. Specifically, given a test mesh $\hat{M}$ and its ground truth mesh $M_{gt}$, \fname{} (\model{}) can be abstracted as 
\begin{equation}
  d = D(\hat{M}, M_{gt}).
  \label{eq:task}
\end{equation}
$d$ is the distance between the test and the ground truth mesh. In this section, we will elaborate on how the distance function $D(\cdot)$ is designed.

\subsection{Overview}
As shown in \cref{fig:pipeline}, our metric is calculated via the following steps: First, we use mesh Fourier transform to analyze the spectrums of the test and ground truth mesh (in \cref{sec:spec}). Then we leverage each frequency band by calculating the Area Under the Curve (AUC) difference of the spectrum curves (in \cref{sec:sad}). Moreover, we further extended our metric by multiplying the AUC difference with a learnable weight to capture the human sensitivity on each frequency band (in \cref{sec:wsad}). We will discuss each step in detail.

\begin{figure*}[t]
\begin{minipage}[h]{0.2\linewidth}
  \centering
   \includegraphics[width=0.82\linewidth]{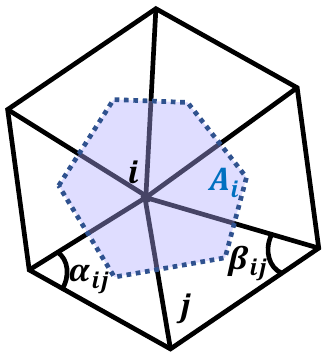}
   \caption{Variables defined in our discrete Laplace-Beltrami operator design.}
   \label{fig:method}
\end{minipage}
\hfill
\begin{minipage}[h]{0.77\linewidth}
  \centering
   \includegraphics[width=1\linewidth]{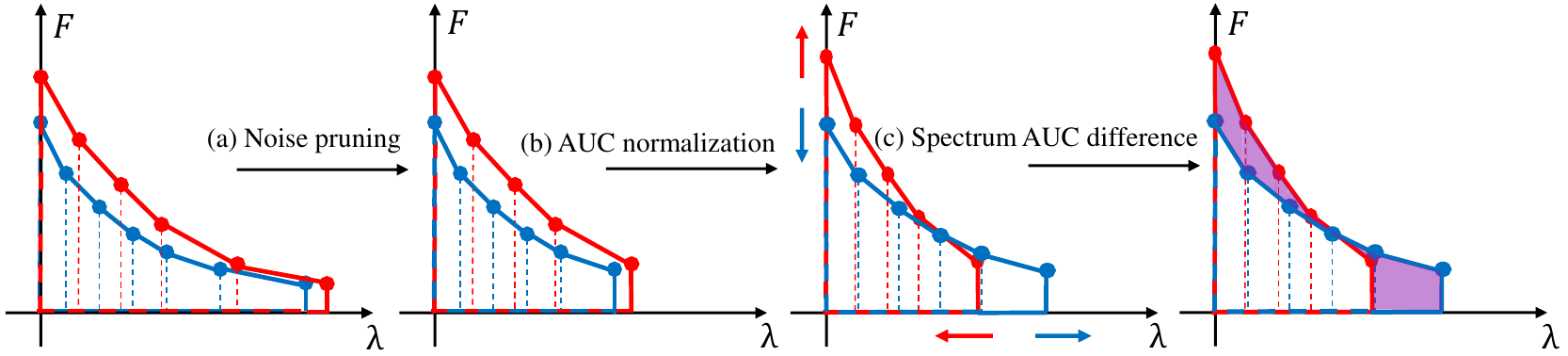}
    \caption{Spectrum Area Under the Curve Difference. We design our metric using the AUC difference of the spectrums. The \textcolor{blue}{blue} curve and \textcolor{red}{red} curve are the test and ground truth mesh spectrum, respectively. The \textcolor{purple}{purple} area in the last graph is the Spectrum AUC Difference. Please find details in \cref{sec:sad}.}
   \label{fig:saucd}
\end{minipage}
\end{figure*}

\subsection{Mesh spectrum analysis}
\label{sec:spec}
In order to capture the overall shape as well as shape details, we choose to decompose the mesh signal into a spectrum. Considering the mesh as a function on a discretized manifold space, we can calculate the spectrum using the manifold space Fourier transform. In Hilbert space, the Fourier operator is defined as the eigenfunctions of the Laplacian operator~\cite{duoandikoetxea2001fourier}. The same definition and similar theories are extended to continuous and discrete manifold space by~\cite{burke1985dg, bobenko2008ddg}. The Laplacian operator on discrete manifold spaces, \ie{} mesh space in our task, is named the Discrete Laplace-Beltrami operator (DLBO). Similar to the Laplacian operator in image space that encodes the pixel information by capturing the local pixel differences~\cite{perez2003poisson,shen2007poisson,wang2007lap,tai2008lap,krishnan2009lap,paris2011lap}, DLBO encodes the mesh shape information by capturing the local shape fluctuation. The ``Cotan formula" defined in~\cite{meyer2003dlbo} is the most widely used discretization, which can be represented in matrix form as
\begin{equation}
\small
    L_{ij}=\left\{
    \begin{array}{cc}
         \sum_{j \in N(i)} \frac{1}{2A_i}(\cot \alpha_{ij} + \cot \beta_{ij}), &  i=j  \\
         -\frac{1}{2A_i} (\cot \alpha_{ij} + \cot \beta_{ij}), &  i \neq j \land j \in N(i)  \\
         0, &  i \neq j \land j \notin N(i),
    \end{array}
    \right.
    \label{eq:cotan}
\end{equation}
where $L \in \mathbb{R}^{N \times N}$ is the DLBO matrix, with $N$ the vertex number of the mesh. $L_{ij}$ indicates its entry in $i$th row and $j$th column, which represents the edge weight between vertex $v_i$ and $v_j$. 
$A_i$ is the mixed Voronoi area of vertex $v_i$ on the mesh. As shown in \cref{fig:method}, the $v_i$'s mixed Voronoi area is defined as the area of the polygon in which the vertices are the circumcenters of $v_i$'s surrounding faces. 
$N(i)$ is the index set of $v_i$'s adjacent vertices. If $v_i$ and $v_j$ are adjacent, $\alpha_{ij}$ and $\beta_{ij}$ are the opposite angles of edge $(v_i, v_j)$ in each of the edge's two neighbor triangle faces, respectively (shown in \cref{fig:method}). If not, $\alpha_{ij}$ and $\beta_{ij}$ are not defined and $L_{ij}$ is 0. As shown in \cref{fig:pipeline}, the DLBO matrix is used for mesh Fourier transform to get mesh spectrum. We calculate the Fourier operator $U^\top$, which is the eigenfunction of $L$ as
\begin{equation}
    L=U\Lambda U^\top,
    \label{eq:eig}
\end{equation}
where $\Lambda$ is a diagonal matrix whose diagonal elements are Fourier mesh frequencies. 

To ensure the mesh frequencies are non-negative, we need the DLBO matrix $L$ to be positive semidefinite. Our experiment in \cref{fig:negative} gives an example of the counterintuitive results when there are negative frequencies. 
However, the Cotan formula in \cref{eq:cotan} does not guarantee to be positive semidefinite. We provide a simple example in \supl{3} in which $L$ is not positive semidefinite when the mesh is not Delaunay triangulated and the mixed Voronoi areas are not all equal to each other. In our metric design, we made two small changes to the original Cotan formula to make it positive semidefinite. a) Inspired by the symmetric normalization of the topology Laplacian matrix in \cite{chung1997spectral}, we make $L$ symmetric by changing the normalization parameter $A_i$ into a symmetric normalized manner $A_i^{\frac{1}{2}}A_j^{\frac{1}{2}}$. b) We replace $\cot \alpha_{ij} + \cot \beta_{ij}$ with $|\cot \alpha_{ij} + \cot \beta_{ij}|$. This ensures all the edge weights in the Laplacian matrix to be non-negative. Thus, our revision of DLBO is defined as
\begin{equation}
\footnotesize
    L_{ij}=\left\{
    \begin{array}{cc}
         \frac{1}{2}\sum_{j \in N(i)} A_i^{-\frac{1}{2}}A_j^{-\frac{1}{2}}|\cot \alpha_{ij} + \cot \beta_{ij}|, &  i=j  \\
         -\frac{1}{2}A_i^{-\frac{1}{2}}A_j^{-\frac{1}{2}} |\cot \alpha_{ij} + \cot \beta_{ij}|, &  i \neq j \land j \in N(i)  \\
         0, &  i \neq j \land j \notin N(i).
    \end{array}
    \right.
    \label{eq:lbo2}
\end{equation}

We prove that our revision of the Cotan formula is positive semidefinite in \supl{B}. In \cref{tab:module}, our experiments show that our DLBO matrix design outperforms the origin Cotan formula in \cite{meyer2003dlbo}, and the topology Laplacian matrix defined in~\cite{chung1997spectral}. 

Finally, we obtain the mesh spectrum by acting Fourier operator on the mesh vertices
\begin{equation}
    F_i=\sqrt{G_{i,x}^2+G_{i,y}^2+G_{i,z}^2}, G=U^\top v,
    \label{eq:fourier}
\end{equation}
where $v \in \mathbb{R}^{N\times 3}$ indicates the 3D coordinates of $N$ mesh vertices. The result spectrum $F\in \mathbb{R}^{N}$. \cref{fig:spec} shows an example of the mesh spectrum (left side) and how the meshes look in different frequency bands (right side). This provides an illustration of the information contained in different frequency bands of the mesh spectrum.

\subsection{Spectrum AUC Difference}
\label{sec:sad}
To reduce the noise and normalize the mesh scale, we also design noise pruning and AUC normalization procedures before calculating the Spectrum AUC Difference.

\textbf{Noise pruning.} 
As shown in \cref{fig:saucd} process \emph{(a)}, we prune a small portion of the highest frequency information to reduce the interference of noise. 
From the observation of the first two meshes (\textcolor{purple}{A} and \textcolor{blue}{B}) in \cref{fig:spec}, we can see that humans can barely notice the shape differences when the highest frequency parts are removed.
Thus, if we try to evaluate the mesh shape that aligns with human perception, it is reasonable to remove high-frequency noise without losing much information that humans care about. Empirically, we choose to prune the highest $0.1\%$ frequency information as noise. Our experiments in \cref{tab:prune} show that this portion is more consistent with human perception while preserving good mesh quality.

\textbf{AUC normalization.} Given a spectrum $F(\lambda)$, its Area Under the Curve (AUC) can be defined as $\int_\infty F(\lambda)d\lambda$.
AUC normalization means using spectrum AUC to normalize the mesh scale. If the mesh scale increases by $s$ times in length, the mixed Voronoi area, \ie{} $A_i$ in \cref{eq:lbo2}, will decrease by $s^2$ times. Thus, each element in the DLBO matrix $L$ will also decrease by $s^2$ times. Then, according to \cref{eq:eig}, the frequency $\lambda$ will decrease $s^2$ times to $\lambda/s^2$, and according to \cref{eq:fourier}, the spectrum amplitude $F$ will change to $sF$ because $v$ is increased by $s$ time and $U^\top$ is still orthonormal.
Then the area under the spectrum curve (the area boxed with \textcolor{red}{red} or \textcolor{blue}{blue} lines in \cref{fig:saucd}) changes as
$A^{\prime} = \int sF(\lambda)d\lambda/s^2 = \frac{1}{s} A$.

In our approach, we normalize the area under the spectrum curve $A$ to 1 to resolve the scale difference, which means $s=A$, $\lambda$ decreases by $A^2$ times, and spectrum amplitude $F$ increases by $A$ times (\cref{fig:saucd} process \emph{(b)}). AUC normalization fairly normalizes the scale of objects in different shapes by only changing the scale, not shape details. It normalizes the spectrum AUC of all mesh to 1, making the mesh spectrums differ only in distributions. Our experiments in \cref{tab:module} demonstrate this design can bring a fairer comparison of the spectrums and improve the human consistency of the metric. The experiment also demonstrates that this design outperforms the spatial domain scale normalization. 

\textbf{Spectrum AUC Difference.} In order to capture the difference between two mesh shapes in the spectrum domain, we design Spectrum AUC Difference (\model{}) on the spectrum analysis results after noise pruning and AUC normalization:
\begin{equation}
    d = D(\hat{M}, M_{gt}) = \int_{\lambda} |\hat{F}(\lambda) - F_{gt}(\lambda)|d\lambda,
    \label{eq:sad}
\end{equation}
where $\hat{F}$ and $F_{gt}$ are the test and groundtruth mesh spectrum. As shown in \cref{fig:saucd} process \emph{(c)}, our metric is defined as the AUC difference of the two spectrum curves (the \textcolor{violet}{purple} area). 
In \cref{tab:module},
we compare our design with an alternative design which changes the amplitude difference $|\hat{F}(\lambda) - F_{gt}(\lambda)|$ to energy difference $|\hat{F}(\lambda)^2 - F_{gt}(\lambda)^2|$. The result shows our design is more consistent with human evaluation. Besides, experiments in \cref{tab:correlation} show our \model{} 
metric aligns well with human evaluation, and outperforms SOTA metrics under multiple evaluation methods. Experiments in \cref{fig:loss} show \model{} has the capability to improve mesh detail qualities in 3D reconstruction when adapted into training loss.

\subsection{Human-adjusted Spectrum AUC Difference}
\label{sec:wsad_sup}
We also provide an extended metric version, in which we design a learnable weight parameter along the frequency bands. The weight parameter indicates the adjustment of human sensitivity to each frequency band. Specifically, we design the extended metric as
\begin{equation}
    d_w = D_w(\hat{M}, M_{gt}) = \int_{\lambda} w(\lambda) |\hat{F}(\lambda) - F_{gt}(\lambda)|d\lambda.
    \label{eq:wsad}
\end{equation}
$w(\lambda)$ is weight parameters indicating human sensitivity along frequency bands. 
Our training loss is defined as
\begin{equation}
    \mathcal{L} = \lambda_{p} \mathcal{L}_{plcc} + \lambda_{sr} \mathcal{L}_{srocc} + \lambda_{r} \mathcal{L}_{regu},
    \label{eq:loss}
\end{equation}
where $\mathcal{L}_{plcc}$ and $\mathcal{L}_{srocc}$ are Pearson correlation loss and Spearman rank order loss. They are defined the same as Pearson's linear correlation~\cite{pearson1920plcc} and Spearman's rank order correlation~\cite{spearman1910srocc}. 
$\mathcal{L}_{regu} = 1/N \sum_i (w_i - 1)^2$ is the regularization loss,
which regularizes weight $w_i$ close to 1. $\lambda_{p}$, $\lambda_{sr}$, and $\lambda_{r}$ are the loss weights of $\mathcal{L}_{plcc}$, $\mathcal{L}_{srocc}$, and $\mathcal{L}_{regu}$. More details of the loss functions can be found in \supl{A}.
Our experiments in \cref{sec:sota} show that after adjustment, the consistency between our metric output and human-annotated ground truth is improved.

\begin{table}
\centering
\resizebox{1\linewidth}{!}{
    \begin{tabular}{|l|c|c|}
    \hline
    dataset & Raw & w/ IQR removal \tabularnewline
    \hline\hline
    number of valid scores & 24304 & 23775 \\
    Scoring range & $[0, 6]$ & $[0, 6]$ \\
    $95\%$ confidence interval & 0.318 & 0.303 \\
    Relative $95\%$ confidence interval & $5.33\%$ & $\textbf{5.04\%}$ \\
    
    \hline
    \end{tabular}
}
\caption{Dataset statistics and error analysis.}
\label{tab:dataset}
\end{table}

\input{tables-metrics}

\section{Experiments}
\label{sec:experiments}

\subsection{Dataset}
We build a user study benchmark dataset \dsname{} to evaluate whether our metric is aligned with human evaluation. 
The dataset contains the human evaluation scores for a variety of distorted meshes. Using this dataset, we can calculate the correlation between metric outputs and human evaluation scores to see how aligned the test metrics are to human evaluation.

\textbf{Dataset design.} We choose 12 objects as ground truth 3D triangle mesh from public object/scene/human mesh datasets such as \cite{Moon_2020data,jensen2014data,wang2022data} and commercial datasets such as \cite{gobo2022data,sketchfab2022data}. These objects are picked from different categories including humans, animals, buildings, plants, \etc{}. For each object, we synthesize 7 different types of distortions which commonly occur in 3D reconstruction. For each distortion type we synthesize 4 distortion levels, which gives us $7 \times 4=28$ distorted objects for every ground truth object. We rotate and render each distorted object into 3 videos using different materials for the mesh. In total, we generate $12 \times 28 \times 3 = 1,008$ distorted mesh videos. \supl{D} shows the meshes and distortion types we use in our dataset along with rendered video examples.

\textbf{Human scoring procedure.} We use a pairwise comparison scoring process similar to~\cite{ponomarenko2009tid2008}. Each subject will evaluate all 28 distorted objects of one ground truth object with a certain material. The scoring follows a Swiss system tournament principle used in~\cite{ponomarenko2009tid2008},
in which each subject takes 6 rounds of pairwise comparison to score the distorted meshes. After 6 rounds of scoring, the meshes are scored from 0 to 6. 0 means the object loses in every round and 6 means it wins in every round. This process will largely reduce the biases among subjects, since the subjects are compelled to distribute an equal amount of points to the 28 distorted objects. The process will take about 15 minutes for each subject, avoiding the fatigue problem in~\cite{bt2002methodology}. For every object rendered with every material, we have 24 to 25 subjects scoring it. In total, we have 868 subjects (536 males, 316 females, and 16 others) who give us $868 \times 28 =24304$ scores. More details of the scoring procedure can be found in \supl{E}.

\textbf{Outlier detection.} We use the interquartile range (IQR) method \cite{dekking2005modern} which is widely used in statistics to detect and remove outliers. For each distorted mesh, we first find the 25 percentile and the 75 percentile of the scores. The score range in between is called the IQR range. We remove the scores that are 1.5IQR smaller than the 25 percentile or 1.5IQR larger than the 75 percentile. Our dataset error analysis in \cref{tab:dataset} shows, that by removing $2.2\%$ of the scores using IQR, we can decrease the uncertainty of the final scoring result by nearly $6\%$.

\textbf{Dataset error analysis.} We analysis the average $95\%$ confidence interval of our dataset scores in~\cref{tab:dataset}. The confidence interval of score $x$ can be calculated as ${\sigma}_{\bar{x}} = z_{0.95} \times \sigma/\sqrt{N}$
where $\sigma$ is the standard derivation of $x$, $N$ is the number of valid scores, and $z_{0.95} \approx  1.96$.
We report the average $95\%$ confidence interval and the relative $95\%$ confidence interval (which is the confidence interval divided by the scoring range). The result shows that dataset scoring is accurate with a $5\%$ error range with IQR outlier removal.

\begin{figure*}[t]
  \centering
   \includegraphics[width=1\linewidth]{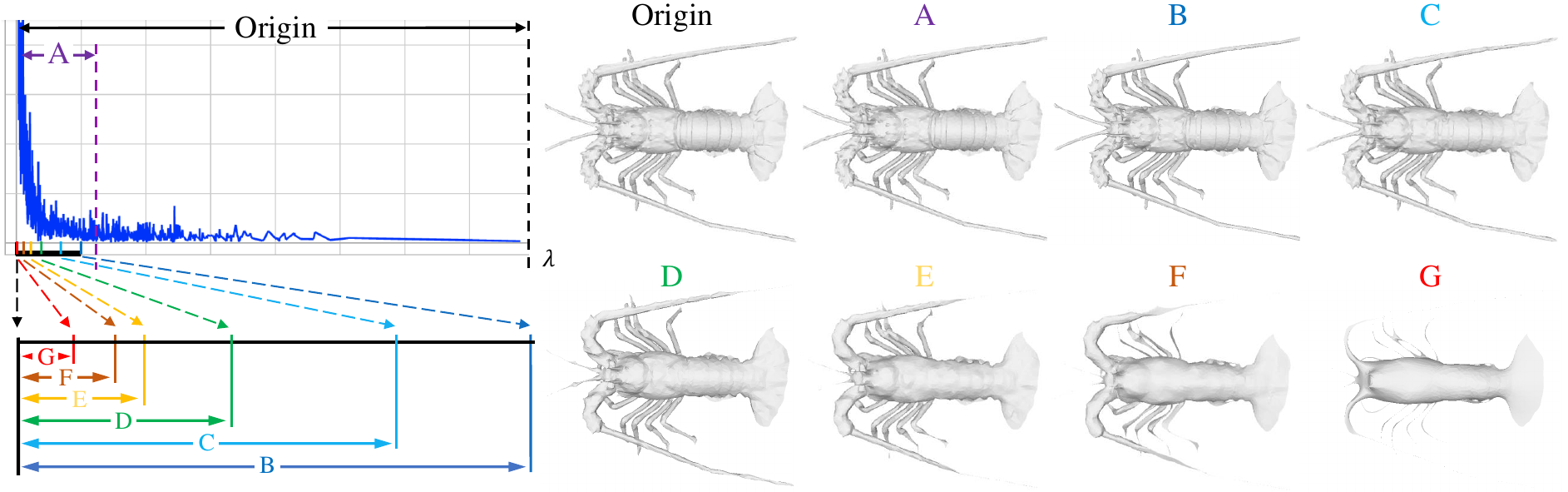}
   \caption{An example of mesh spectrum curve: We do mesh Fourier transform on the ``Origin" mesh and show the spectrum in the left graph. The $\lambda$-axis is the eigenvalues of the DLBO matrix, the larger the higher frequency. We also show how mesh changes when gradually removing high-frequency information (mesh \textcolor{violet}{A} to \textcolor{red}{G}). The frequency bands of the meshes are shown as the colored arrows in the left graph.}
   \label{fig:spec}
\end{figure*}

\textbf{Evaluation methods.}
We use 3 different evaluation methods to evaluate the correlation between our metrics and the human scoring (ground truth) on our \dsname{} benchmark dataset. Pearson's linear correlation coefficient (\textbf{PLCC})~\cite{pearson1920plcc} is used to evaluate the linear alignment between our metric and human perception. We also used Spearman's rank order correlation coefficient (\textbf{SROCC})~\cite{spearman1910srocc} and Kendall's rank order correlation coefficient (\textbf{KROCC})~\cite{kendall1946advanced} to evaluate the ranking order correlation between our metric and human perception. The possible ranges of 3 metrics are all $[-1, 1]$. Higher numbers mean stronger correlations. More details of the three evaluation methods can be found in \supl{A}.

\subsection{Implementation details}
\label{sec:imple}
We implement our basic version metric following \cref{eq:sad}. 
$\hat{F}(\lambda)$ and $F_{gt}(\lambda)$ in \cref{eq:sad} are both piece-wise functions, so we implement the integration by simply adding every piece area together.
We implement our human-adjusted version following \cref{eq:wsad}.  We use a 20-dimensional weight $w(\lambda)$ to avoid overfitting. We interpolate $w$ to all frequencies of the ground truth and test meshes and element-wisely multiply them to the spectrums. In spectrum weight training, SROCC and PLCC are used as part of the loss function as \cref{eq:loss}. KROCC is not used in training but only for testing. We use a k-fold strategy for training the human-adjusted weight. Each time we choose 1 object for testing and the rest 11 objects for training, which means $k=12$. More implementation details can be found in \supl{A}.

\subsection{Quantitive and qualitative results}
\label{sec:sota}

\textbf{SOTA comparison.} \cref{tab:correlation} shows our results compared to previous 3D mesh shape metrics. We evaluated the correlation between each metric and the human scoring via three different evaluation methods. We observe that \textbf{a)} without any learning-based design, our metric outperforms the SOTA learning-based (SSFID) and non-learning-based metrics (Chamfer Distance, IoU, F-score, and UHD), \textbf{b)} our extended version metric with learned weights has better linearity and slightly better ranking order correction with human evaluation, and \textbf{c)} our results on different objects show that our metrics have good generalizability.

\textbf{Spectrum example.} We first show an example of mesh spectrum in \cref{fig:spec}. We decompose the ``origin" mesh using the Fourier Transform and get the resulting spectrum (top-left graph). The meshes on the right (from mesh \textcolor{violet}{A} to \textcolor{red}{G}) are generated by gradually removing high-frequency information. The frequency bands of the meshes are shown as colored arrows in and under the graph. As we see, the details gradually disappear as we remove high-frequency information.

\textbf{Frequency band separation.} We explored the consistency between human perception and the information obtained from every frequency band. Specifically, we separate the frequency band exponentially and build metrics only using information from that frequency band. The results are shown in \cref{tab:freqband}, we find the frequency bands $[0, 0.001]$ and $[0.01, 0.03]$ have the best consistency with human perception. Moreover, it shows that if we put all frequencies together, they can achieve better results.

\begin{figure}[t]
  \centering
   \includegraphics[width=1\linewidth]{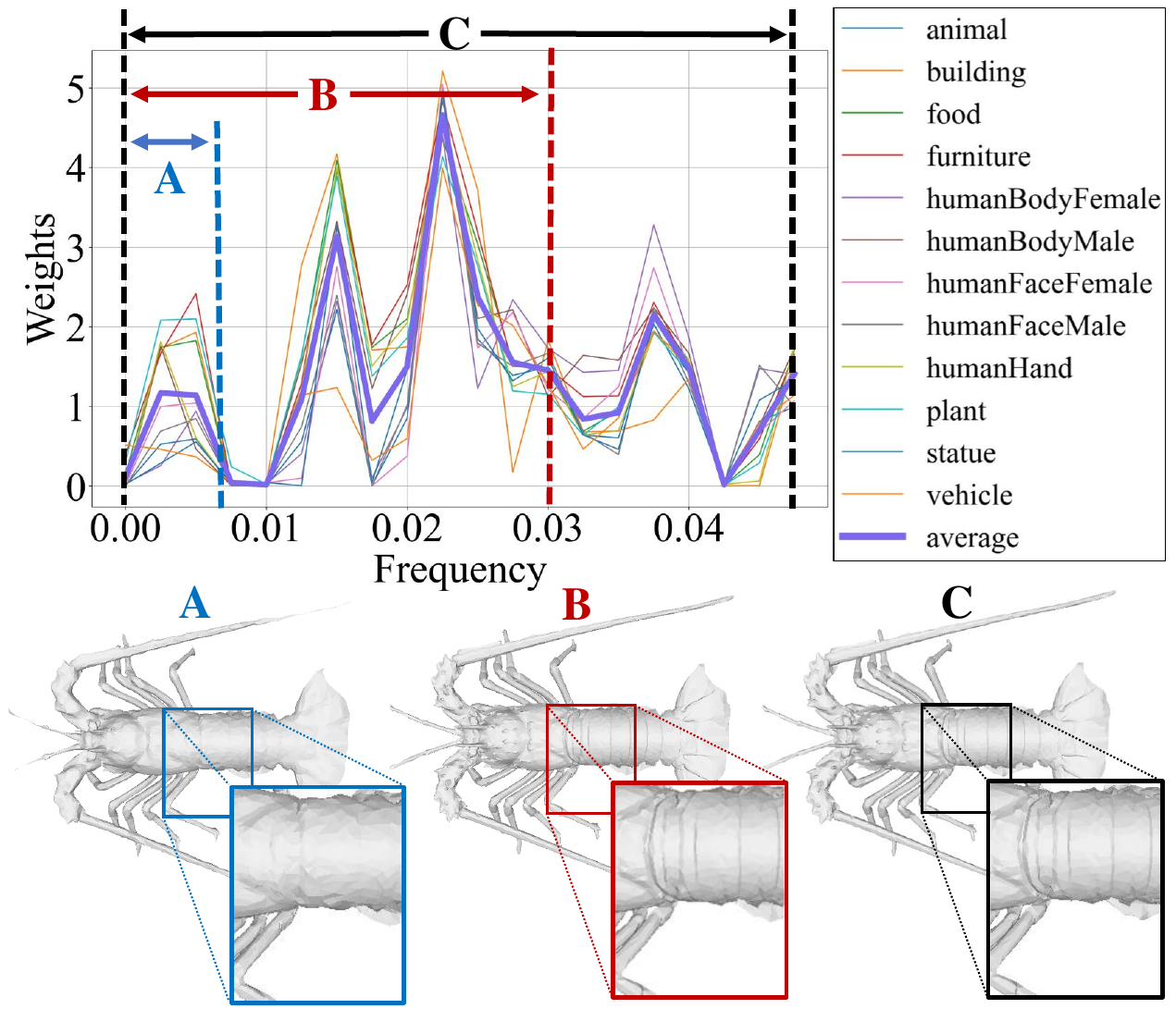}

   \caption{Learned spectrum weights on all 12 folds. The name of colorful thin lines means the test object name of that fold. The bold \textbf{\textcolor{violet}{purple}} line is the average weights of all folds. We also show some examples of mesh shape information in different frequency bands. Frequency band \textcolor{blue}{A} is $[0, 0.0075)$, \textcolor{red}{B} is $[0, 0.03)$, and \textcolor{black}{C} is $[0, 0.05)$.}
   \label{fig:weights}
\end{figure}

\begin{figure*}[t]
  \centering
   \includegraphics[width=1\linewidth]{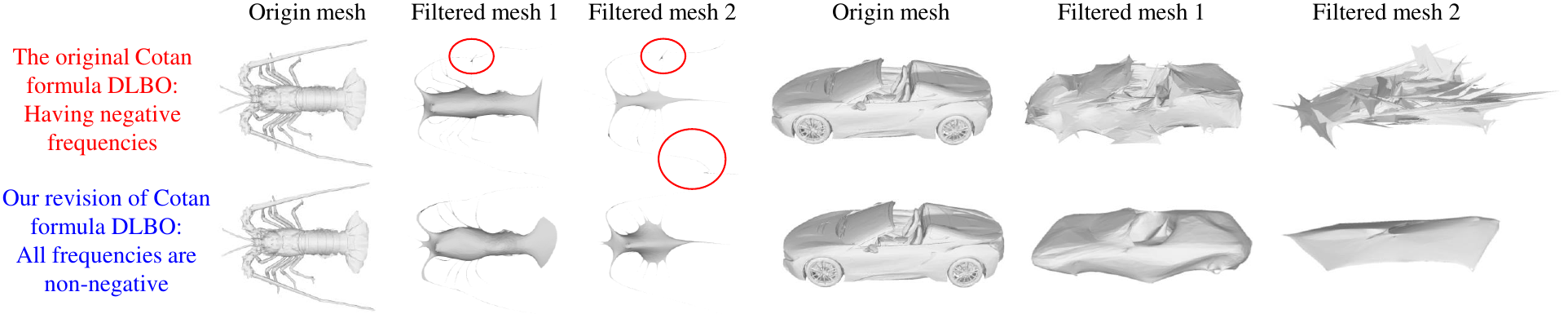}
   \caption{Counterintuitive low-frequencies information if some of the mesh frequencies are negative. We can see if we remove the high-frequency part of the mesh (resulting in ``Filtered mesh 1" and ``Filtered mesh 2") using the original Cotan formula, the mesh's low-frequency parts show artifacts (sharp shapes). The \textcolor{red}{red} circles show the artifacts in the left object. The right object shows a case when these artifacts occur much more often. These artifacts do not occur using our revised Cotan formula DLBO.}
   \label{fig:negative}
\end{figure*}

\input{tables-ablation.tex}

\begin{figure}[t]
  \centering
   \includegraphics[width=1\linewidth]{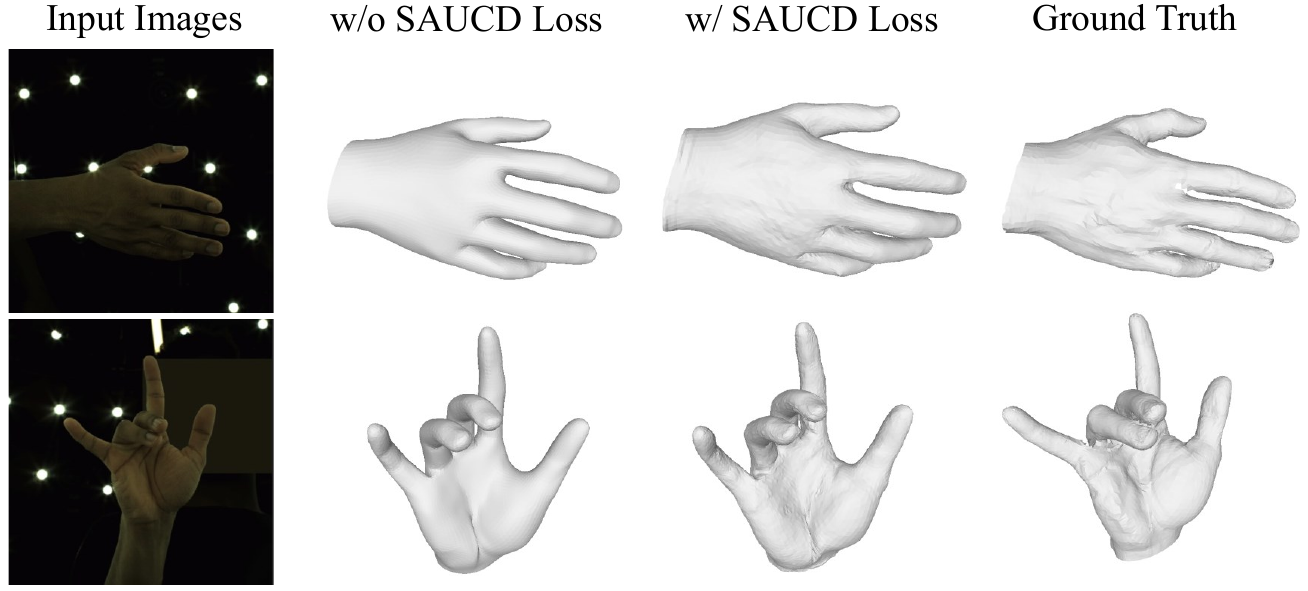}
   \caption{We Adapt \model{} into a loss function and use it in monocular-image-based 3D hand reconstruction. From left to right: input images, reconstruction result w/o \model{} loss, reconstruction result w/ \model{} loss, and ground truth mesh. We can see that the enhancement of \model{} loss in mesh details is clearly noticeable.}
   \label{fig:loss}
\end{figure}

\textbf{Trained weight.} We show our trained weights in Human-adjusted metric in \cref{fig:weights}. Different lines represent different folds, and the bold \textbf{\textcolor{violet}{purple}} line is the average weight. We can see the weights trained on each fold have similar patterns. We also observe that the weight curves have a small peak in the range \textcolor{blue}{A} and two much larger peaks between \textcolor{blue}{A} and \textcolor{red}{B}, which means our extended metric relies more on the information between \textcolor{blue}{A} and \textcolor{red}{B}. We show an example of mesh shapes in the range \textcolor{blue}{A}, \textcolor{red}{B}, and \textcolor{black}{\textbf{C}} at the bottom of \cref{fig:weights}. Mesh \textcolor{blue}{A} obviously has fewer details than Mesh \textcolor{red}{B}, and the weight curve shows that this difference is what the learning process tries to emphasize.

\textbf{Negative frequencies.} In \cref{fig:negative} 
we illustrate how our revised Cotan formula DLBO in \cref{eq:lbo2} improves frequency analysis compared to the original Cotan formula in \cref{eq:cotan}~\cite{meyer2003dlbo}.
The first and second rows are the results of the original Cotan formula DLBO and our revised Cotan formula DLBO, respectively. The original Cotan formula can yield negative frequencies due to its lack of positive semidefiniteness, whereas our revision ensures all frequencies are non-negative. For both objects in the figure, we remove different portions of high-frequency information and show the remaining low-frequency parts (resulting in 
``Filtered mesh 1" and ``Filtered mesh 2"). For the left object, notice the counterintuitive sharp shapes in the \textcolor{red}{red} circle when using the Cotan formula. The right object is a much more severe case. Sharp shapes in low-frequency parts show improper decomposition and high-frequency aliasing with low-frequency shapes, making the Cotan formula unsuitable for spectral mesh comparison. In contrast, our revised formula yields smooth low-frequency components without these artifacts.

\textbf{Noise pruning portion.} \cref{tab:prune} shows our \model{} metric performance by changing the noise pruning portion (\cref{sec:sad}). The metric achieves better results when the pruning portion is $0.1\%$ or $1\%$. In our proposed metric, we choose the pruning portion to be $0.1\%$ to best avoid possible high-frequency information loss.

\textbf{Module replacement.} \cref{tab:module} shows our \model{} metric performance by replacing some modules with alternative designs. First, we replace our revision of the discrete Laplace-Beltrami operator in \cref{eq:lbo2} with topology Laplacian matrix in~\cite{chung1997spectral} and ``Cotan formula" in~\cite{meyer2003dlbo}.
Second, we change the AUC difference defined in \cref{eq:sad} into the energy difference, which means changing $|\hat{F}(\lambda) - F_{gt}(\lambda)|$ in \cref{eq:sad}
into $|\hat{F}(\lambda)^2 - F_{gt}(\lambda)^2|$.
In the third experiment, we replace AUC normalization (in \cref{sec:sad}) with spatial normalization, where we normalize the meshes by their maximum range along all 3 spatial axes. We also removed the AUC normalization module for another comparison. Our experiments show \model{} has better performance than alternative designs.

\textbf{Adapting \model{} to loss function.} We adapted our metric into a loss function to enhance the visual quality of 3D mesh reconstructions, as evident from the hand reconstruction results in \cref{fig:loss}. Details on the experiment's implementation are available in \supl{G}. From this experiment, we can see that the enhancement of \model{} loss in mesh details is clearly noticeable.

\textbf{Visualized examples.} We visualize examples in our dataset and their evaluation result using different metrics in \supl{F}.

\section{Conclusions}
\label{sec:conclusion}
In order to propose a 3D shape evaluation that better aligns with human perception, we design an analytic metric named \fname{} (\model{}).
Our proposed \model{} leverages mesh spectrum analysis 
to evaluate 3D shape that aligns with human evaluation, and its extended version Human-adjusted \model{} further explores the sensitivity of human perception of each frequency band. 
To evaluate our new metrics, we build a user study dataset to compare our metrics with existing metrics. The results validate that both our new metrics are well aligned with human perceptions and outperform previous methods. 


\appendix
\section*{Appendix}
\section{Implementation Details}
\label{sec:imple_sup}

\subsection{Discretization of \fname{}}

Our \fname{} (\model{}) is defined in \meq{7} as
\begin{equation}
    d = D(\hat{M}, M_{gt}) = \int_{\lambda} |\hat{F}(\lambda) - F_{gt}(\lambda)|d\lambda,
    \label{eq:sad_sup}
\end{equation}
where $\hat{F}(\lambda)$ and $F_{gt}(\lambda)$ are the test and groundtruth mesh spectrum, respectively. To discretize \cref{eq:sad_sup} in the experiments, we let $\{\hat{\lambda}_i\}$ to be the discretized frequencies of $\hat{F}(\lambda)$ and $\{\lambda_{gt,i}\}$ to be the discretized frequencies of $F_{gt}(\lambda)$. We sort the two sets $\{\hat{\lambda}_i\}$ and $\{\lambda_{gt,i}\}$ into one array from low to high, resulting in a sorted array $\{\lambda_i\}$ with $N_{gt} + \hat{N}$ frequencies, where $N_{gt}$ is the vertex number of the ground truth mesh and $\hat{N}$ is the vertex number of the test mesh. The $N_{gt} + \hat{N}$ frequencies discretize \cref{eq:sad_sup} into the sum of the area of $N_{gt} + \hat{N}- 1$ segments as: 
\begin{equation}
    d = \sum_{i=1}^{N_{gt} + \hat{N}-1} s_i,
    \label{eq:dsad}
\end{equation}
where the area of each segment 
\begin{equation}
    s_i = \left\{
    \begin{array}{cc}
         \frac{1}{2}|H_i+H_{i-1}|(\lambda_i - \lambda_{i-1}), & H_iH_{i-1}\geq 0\\
         \frac{H_i^2+H_{i-1}^2}{2|H_i+H_{i-1}|}(\lambda_i - \lambda_{i-1}), & H_iH_{i-1}< 0,
    \end{array}
    \right.
\label{eq:si}
\end{equation}
is either a trapezoid when $H_iH_{i-1}\geq 0$ or two triangles when $H_iH_{i-1}< 0$. Here, 
\begin{equation}
H_i = \hat{F}(\lambda_i) - F_{gt}(\lambda_{i})
\label{eq:Hi}
\end{equation}
is the amplitude difference between $\hat{F}(\lambda)$ and $F_{gt}(\lambda)$ at $\lambda_i$.
If $\lambda_i$ is originally from the test mesh spectrum, then 
\begin{equation}
\hat{F}(\lambda_i) = \hat{F}(\hat{\lambda}_i),
\label{eq:nointer}
\end{equation}
and $F_{gt}(\lambda_{i})$ is calculated using interpolation as
\begin{equation}
\small
F_{gt}(\lambda_{i}) = \frac{(\lambda_{gt, i+} - \lambda_{i})F_{gt}(\lambda_{gt, i+}) + (\lambda_{i} - \lambda_{gt, i-})F_{gt}(\lambda_{gt, i-})}{\lambda_{gt, i+} - \lambda_{gt, i-} },
\label{eq:inter}
\end{equation}
where $\lambda_{gt, i-}$ and $\lambda_{gt, i+}$ are the left and right nearest frequencies of $\lambda_i$ in the groundtruch frequency set $\{\lambda_{gt,i}\}$. 
Similarly, if $\lambda_i$ is originally from the ground truth mesh spectrum, then 
\begin{equation}
F_{gt}(\lambda_i) = F_{gt}(\lambda_{gt,i}),
\label{eq:nointer2}
\end{equation}
and $\hat{F}(\lambda_{i})$ is calculated using interpolation as
\begin{equation}
\hat{F}(\lambda_{i}) = \frac{(\hat{\lambda}_{i+} - \lambda_{i})\hat{F}(\hat{\lambda}_{i+}) + (\lambda_{i} - \hat{\lambda}_{i-})\hat{F}(\hat{\lambda}_{i-})}{\hat{\lambda}_{i+} - \hat{\lambda}_{i-} },
\label{eq:inter2}
\end{equation}
where $\hat{\lambda}_{i-}$ and $\hat{\lambda}_{i+}$ are the left and right nearest frequencies of $\lambda_i$ in the test frequency set $\{\hat{\lambda}_i\}$.

In summary, to calculate the area of the region between the two curves (\ie{} AUC difference), we first sort the frequencies from the test and ground truth spectrum in one array, and interpolate the test and ground truth spectrum using the frequencies from the other spectrum. Then, we calculate each AUC difference in the range between two adjacent frequencies and add them together. When $H_iH_{i-1}\geq 0$, the region between the two curves is a trapezoid; when $H_iH_{i-1}< 0$ the region is two triangles and we calculate the sum area of the two triangles. Finally, the sum of the areas between adjacent frequencies is our \fname{} metric.

\subsection{Discretization of Human-adjusted \model{}}

\label{sec:wsad}
Our Human-adjusted \model{} is defined in \meq{8} as
\begin{equation}
    d = D(\hat{M}, M_{gt}) = \int_{\lambda} w(\lambda) |\hat{F}(\lambda) - F_{gt}(\lambda)|d\lambda.
    \label{eq:wsad_sup}
\end{equation}
Similar to SAUCD discretization, Human-adjusted SAUCD can be discretized as
\begin{equation}
    d = \sum_{i=1}^{N_{gt} + \hat{N}-1} w_is_i,
    \label{eq:dwsad}
\end{equation}
where $s_i$ is defined the same as in \cref{eq:dsad}, and $w_i$ is the human-adjusted weight at $\lambda_i$ in \cref{eq:si}. Since the weight vector $\mathbf{w}$ we use is only 20-dimensional to avoid overfitting, we get each $w_i$ by interpolating $\mathbf{w}$ at each $\lambda_i$. Specifically, the 20 elements of $\mathbf{w}$ represent the weights at frequencies uniformly distributed in the range from 0 to 0.05. We denote those 20 frequencies as $\{\lambda_{\mathbf{w}, k}\}$ on which the weights $\mathbf{w}$ are explicitly defined, which means $0 \leq k< 20$, $\lambda_{\mathbf{w},0} = 0$, and $\lambda_{\mathbf{w},19} = 0.05$. The last frequency location 0.05 is picked empirically. Note that we use a revised version of Discrete Laplace-Beltrami Operator (DLBO) as in \meq{4} to make sure $\lambda_i \geq 0$, then to calculate weight $w_i$ whose corresponding $\lambda_i \notin \{\lambda_{\mathbf{w}, k}\}$, we only consider when $\lambda_i>0$. We use interpolation to calculate $\lambda_i$ as
\begin{equation}
\small
    w_i = \left\{
    \begin{array}{cc}
        \frac{(\lambda_{\mathbf{w},i+} - \lambda_{i})\mathbf{w}(\lambda_{\mathbf{w},i+}) + (\lambda_{i} - \lambda_{\mathbf{w}, i-})\mathbf{w}(\lambda_{\mathbf{w},i-})}{\lambda_{\mathbf{w}, i+} - \lambda_{\mathbf{w}, i-}}, & 0<\lambda_i < \lambda_{\mathbf{w},19} \\
        \lambda_{\mathbf{w},19}, & \lambda_i > \lambda_{\mathbf{w},19},
    \end{array}
    \right.
    \label{eq:interw}
\end{equation}
where $\lambda_{\mathbf{w},i-}$ and $\lambda_{\mathbf{w}, i+}$ are the left and right nearest element to $\lambda_i$ in $\{\lambda_{\mathbf{w}, k}\}$. 

Having $w_i$, we can calculate Human-adjusted \model{} following \cref{eq:dwsad}.

\subsection{Evaluation methods}
\label{sec:eval}
We use 3 different evaluation methods to evaluate the correlation between our metrics and human scoring (ground truth) on our provided \dsname{} dataset.

\textbf{Pearson's linear correlation coefficient (PLCC).} Pearson's correlation~\cite{pearson1920plcc} evaluates the linear alignment between our metrics and human evaluation. It is defined as
\begin{equation}
    p =\frac{\sum_{i=1}^N(h_i-\bar{h}_i)(m_i-\bar{m}_i)}{\sqrt{\sum_{i=1}^N(m_i-\bar{m}_i)^2}\sqrt{\sum_{i=1}^N(h_i-\bar{h}_i)^2}},
    \label{eq:plcc}
\end{equation}
where $m_i$ is the score of mesh $i$ given by the tested metric and $h_i$ is the groundtruth score (human scoring) of mesh $i$. $\bar{h}_i$ and $\bar{m}_i$ are the average score of $h_i$ and $m_i$, respectively.

\textbf{Spearman's rank order correlation coefficient (SROCC).} SROCC~\cite{spearman1910srocc} is one of the most commonly used metrics to measure rank correlations. It is defined as
\begin{equation}
    r_s=1-\frac{6\sum (R(m_i)-R(h_i))^2}{n(n^2-1)},
    \label{eq:srocc}
\end{equation}
where $m_i$ and $h_i$ is are defined the same as in \cref{eq:plcc}. $R(m_i)$ and $R(h_i)$ are the rankings of $m_i$ and $h_i$, and $n$ is the amount of data. In our paper, $n$ is the number of meshes scored by one subject.

\textbf{Kendall's rank order correlation coefficient (KROCC).} KROCC~\cite{kendall1946advanced} is also a rank order correlation. It is defined as
\begin{equation}
    \tau =1-\frac{2}{n(n^2-1)} \sum_{i<j} sgn(m_i-m_j)sgn(h_i-h_j),
    \label{eq:krocc}
\end{equation}
where $m_i$, $h_i$, and $n$ is the same with \cref{eq:srocc}, and $sgn(\dot)$ is the sign function, which means $sgn(x) = 1$ when $x>0$, $sgn(x) = -1$ when $x<0$, and $sgn(x) = 0$ when $x=0$. The difference between SROCC and KROCC is that SROCC considers the actual amount of rank order difference of input data, while KROCC only counts the number of inverse pairs.

The possible ranges of all 3 metrics are $[-1, 1]$. Higher numbers mean stronger correlations.

\subsection{Human-adjusted SAUCD training}
During training, Pearson's correlation loss $\mathcal{L}_{plcc}$ and Spearsman's rank order loss $\mathcal{L}_{srocc}$ in \meq{9} are defined the same as \cref{eq:plcc} and \cref{eq:srocc}, respectively. Note that, since the rank part of SROCC is not naturally differentiable, we used a differentiable ranking approach provided in~\cite{blondel2020softrank} to make \cref{eq:srocc} differentiable. We set $\lambda_{p}=0.1$ , $\lambda_{sr}=10$, and $\lambda_{regu}=1$ for \meq{9}. The training process took about 1 minute on a 14-core Intel Xeon CPU. The training code is implemented using PyTorch~\cite{paszke2019pytorch}.

\section{Proof of Positive-semidefiniteness of Revised Cotan Formula}
In this section, we prove that our revised version of the Cotan formula in \meq{4} is positive semidefinite. Here, the DLBO defined in \meq{4} is
\begin{equation}
\footnotesize
    L_{ij}=\left\{
    \begin{array}{cc}
         \frac{1}{2}\sum_{j \in N(i)} A_i^{-\frac{1}{2}}A_j^{-\frac{1}{2}}|\cot \alpha_{ij} + \cot \beta_{ij}|, &  i=j  \\
         -\frac{1}{2}A_i^{-\frac{1}{2}}A_j^{-\frac{1}{2}} |\cot \alpha_{ij} + \cot \beta_{ij}|, &  i \neq j \land j \in N(i)  \\
         0, &  i \neq j \land j \notin N(i).
    \end{array}
    \right.
    \label{eq:lbo2_sup}
\end{equation}
According to the Gershgorin circle theorem~\cite{horn2012matrix}, for every eigenvalue $\lambda_k$ of $L$, 
\begin{equation}
\lambda_k \in \bigcup_i S_i,
\label{eq:lambdak}
\end{equation}
where $S_i$ is the $i$th Gershgorin disc. The Gershgorin disc is defined as 
\begin{equation}
S_i = \{z\in \mathbb{C}:|z-L_{ii}| \leq R_i = \sum_{i\neq j}|L_{ij}|\},
\end{equation}
where $\mathbb{C}$ means the complex space. Since $L$ is a real symmetric matrix, according to \cref{eq:lbo2_sup}, the Gershgorin disc degenerates into a line segment in the real space as
\begin{equation}
S_i = \{s\in \mathbb{R}:|s-L_{ii}| \leq R_i = \sum_{i\neq j}|L_{ij}|\}.
\label{eq:Si}
\end{equation}
From ~\cref{eq:lbo2_sup}, we can also have
\begin{equation}
 \sum_{i\neq j}|L_{ij}| = \sum_{j \in N(i)} \frac{| \cot \alpha_{ij} + \cot \beta_{ij}|}{2\sqrt{A_iA_j}} = L_{ii}.
 \label{eq:Lii}
\end{equation}
Note that $L_{ii} \geq 0$, so having \cref{eq:Lii}, from \cref{eq:Si} we get 
\begin{equation}
S_i = \{s\in R:|s-L_{ii}| \leq R_i = L_{ii}\}
\Leftrightarrow 0 \leq S_i \leq 2L_{ii}. 
\end{equation}
Thus, according to \cref{eq:lambdak}, we have 
\begin{equation}
0 \leq \lambda_k \leq 2 \max_i L_{ii}, \forall 0\leq k \leq N,
\end{equation}
where $N$ is the number of vertices. Then, $L$ is positive semidefinite since $L$ is a real symmetric matrix and all its eigenvalues are greater than or equal to zero.

Q.E.D.

\begin{figure}[t]
  \centering
   \includegraphics[width=1\linewidth]{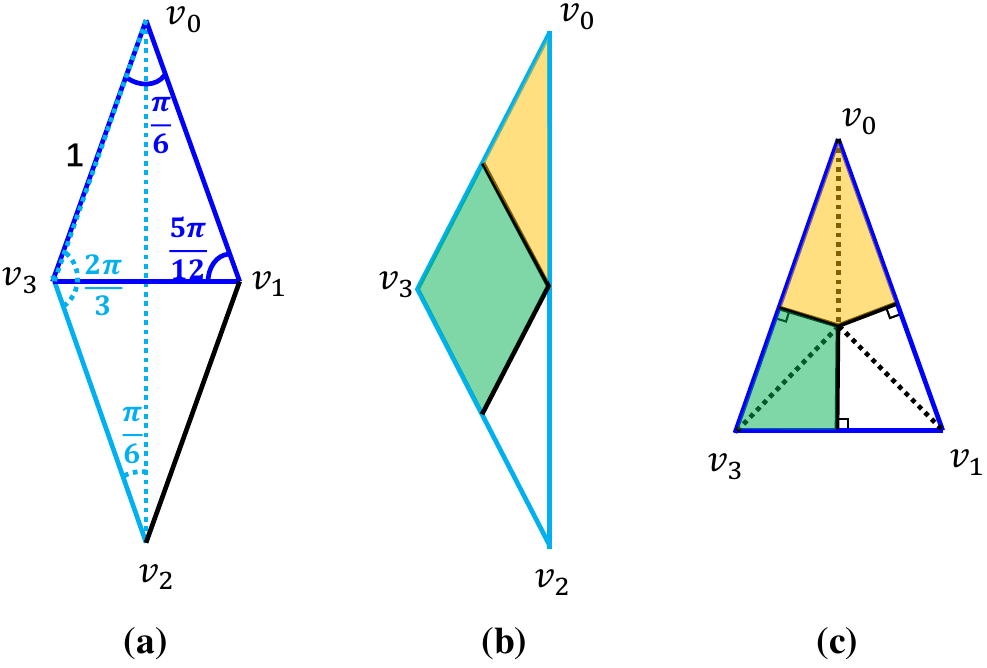}
   \caption{A simple mesh example to show that the original Cotan formula does not guarantee to be positive semidefinite.}
   \label{fig:example}
\end{figure}

\begin{table*}[t]
  \centering
    \begin{tabular}{|p{2cm}|p{3cm}|p{11.5cm}|}
    \hline
    \rowcolor{red!20}Distortion types & Description & Generating details \tabularnewline
    \hline\hline
    Impulse & Adding impulsive noise on mesh surface & We add Gaussian noise on $r$ percent of the ground truth mesh vertices. The mean of the Gaussian noise is set to 0 and standard derivation is set to $\sigma$ percent of the mesh scale. For 4 levels of this distortion, ($r$, $\sigma$) are set to (1, 0.5), (5, 2), (8, 3), and (1, 5), respectively. \\
    \rowcolor{red!10}Poisson reconstruction noise & Synthesizing the noise occurs in Poisson reconstruction~\cite{kazhdan2006poisson} & We first use Poisson disk sampling~\cite{bridson2007poissondisk} to sample $sN$ points from the groundtruth mesh surface, where $N$ is the number of vertices in groundtruth mesh. Then, we use Poisson reconstruction provided in MeshLab~\cite{cignoni2008meshlab} to reconstruct the mesh surface from the sampled points. The reconstruction depth is set to 6. For 4 levels of this distortion, $s$ is set to 0.9, 0.5, 0.2, and 0.05, respectively.\\
    Smoothing & Smoothing mesh surface& We apply $i$ times of $\lambda-\mu$ Taubin smoothing~\cite{taubin1995curve} to smooth the groundtruth mesh surface, where $\lambda=0.5$ and $\mu=-0.53$.  For 4 levels of this distortion, $i$ is set to 5, 20, 50, and 200, respectively.\\
    \rowcolor{red!10}Unproportional scaling & Stretching (or shrinking) the mesh along $x$, $y$, and $z$ axis with different rates& We stretch the mesh to $s_x$ percent to its original length along $x$ axis, and shrink the mesh to $s_z$ percent to its original length along $z$ axis. For 4 levels of this distortion, ($s_x$, $s_z$) are set to (98, 102), (95, 105), (90, 110), (80, 120), respectively. \\
    Low-resolution mesh &Simplifying mesh surface to lower resolution& We simply the ground truth mesh surface using edge collapse algorithm~\cite{garland1997surface}. For 4 levels of this distortion, the target face number is set to 5000, 2000, 1000, and 500, respectively.\\
    \rowcolor{red!10}White noise & Adding Gaussian white noise on mesh surface& We add Gaussian noise on \emph{all} the groundtruth mesh vertices. The mean of the Gaussian noise is set to 0 and standard derivation is set to $\sigma$ percent of the mesh scale. For 4 levels of this distortion, $\sigma$ is set to 0.1, 0.2, 0.3, and 0.5, respectively. \\
    Outlying noise & Adding outlying small floating spheres around the mesh& We add floating spheres around the ground truth mesh to synthesize outlying noise that occurs in 3D reconstruction. The number of the spheres is set to $n$ and the radius $rA$, where $A$ is the maximum length of the mesh along $x$, $y$, and $z$ dimensions. The locations of the spheres are sampled randomly from a cube that surrounds the ground truth mesh. The edge size of the cube is set to $(1 + 6r)A$. For 4 levels of this distortion, ($n$, $r$) are set to (20, 0.002), (30, 0.004), (40, 0.006), (80, 0.008), respectively.\\
    \hline
    \end{tabular}
   \caption{Distortions in our provided \emph{Shape Grading} dataset.}
   \label{tab:distortion}
\end{table*}

\begin{figure*}[t]
  \centering
   \includegraphics[width=1\linewidth]{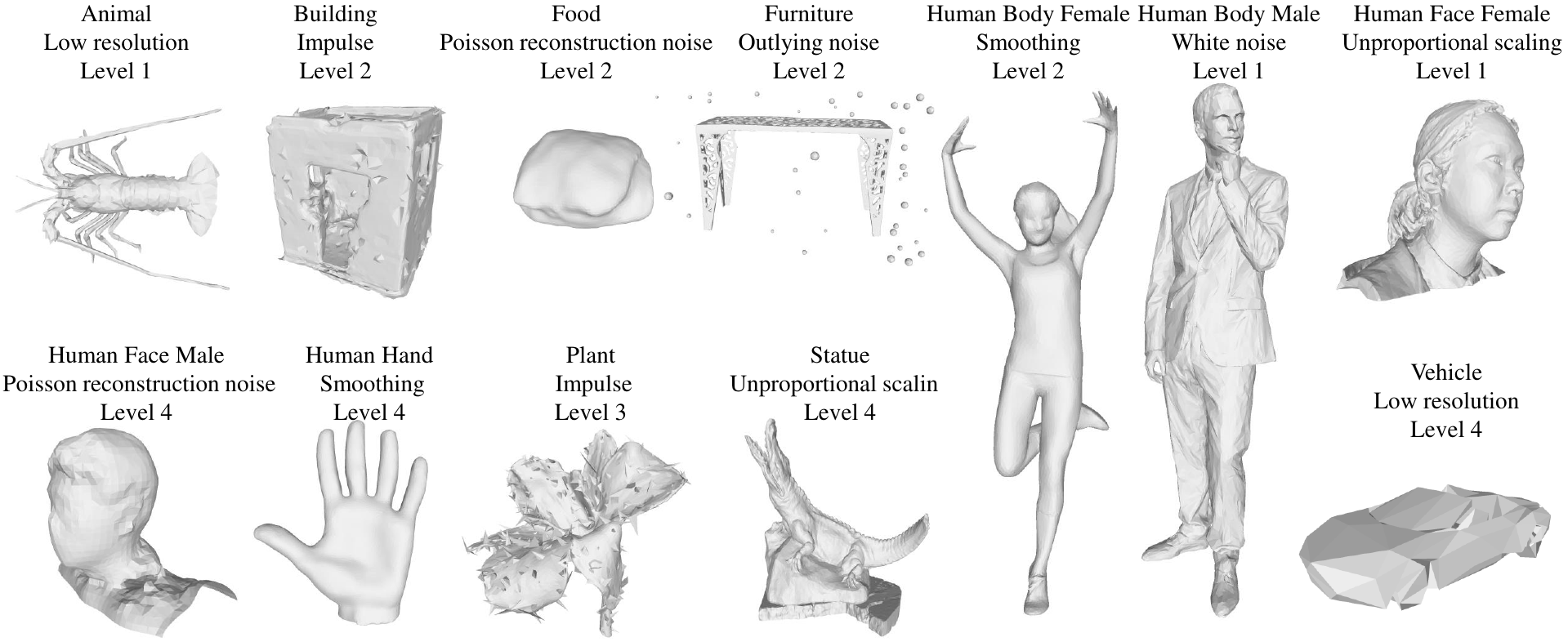}
   \caption{Examples of distorted meshes of different distortion levels in our provided \emph{Shape Grading} dataset.}
   \label{fig:distortion}
\end{figure*}

\begin{figure*}[t]
  \centering
   \includegraphics[width=1\linewidth]{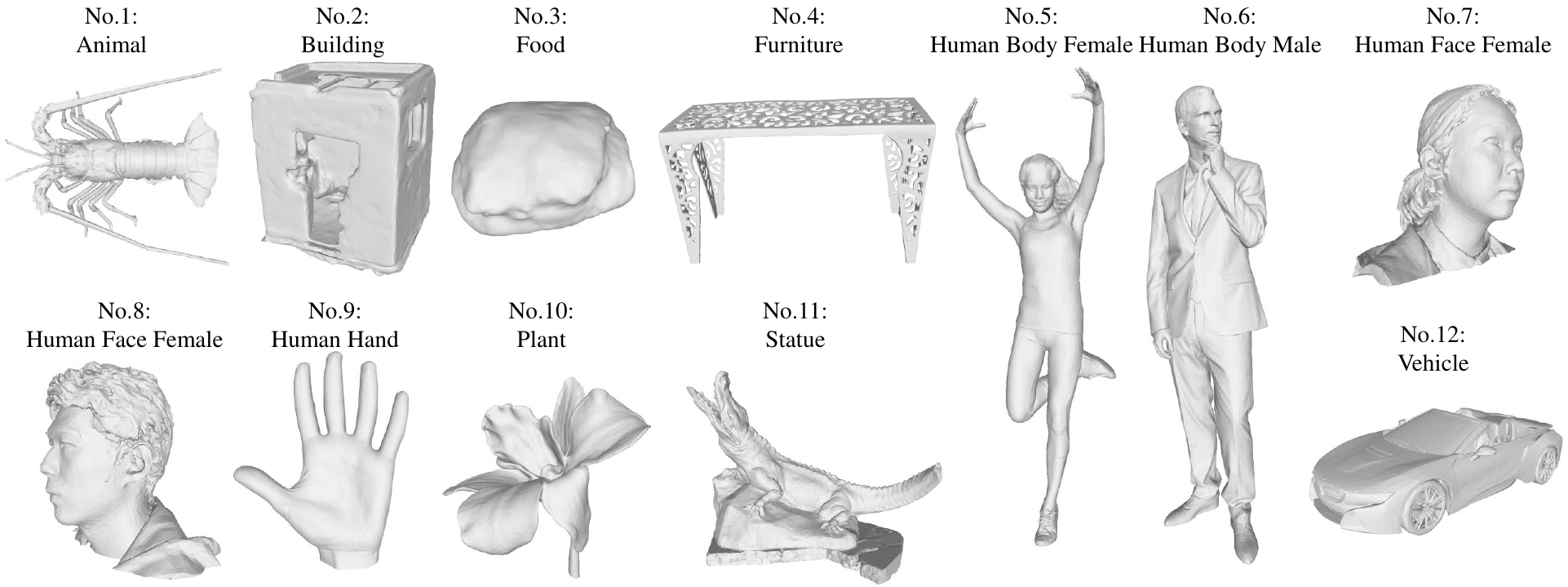}
   \caption{Objects in our provided \emph{Shape Grading} dataset and what the object numbers correspond to in \mtab{2}.}
   \label{fig:object}
\end{figure*}

\begin{figure*}[t]
  \centering
   \includegraphics[width=1\linewidth]{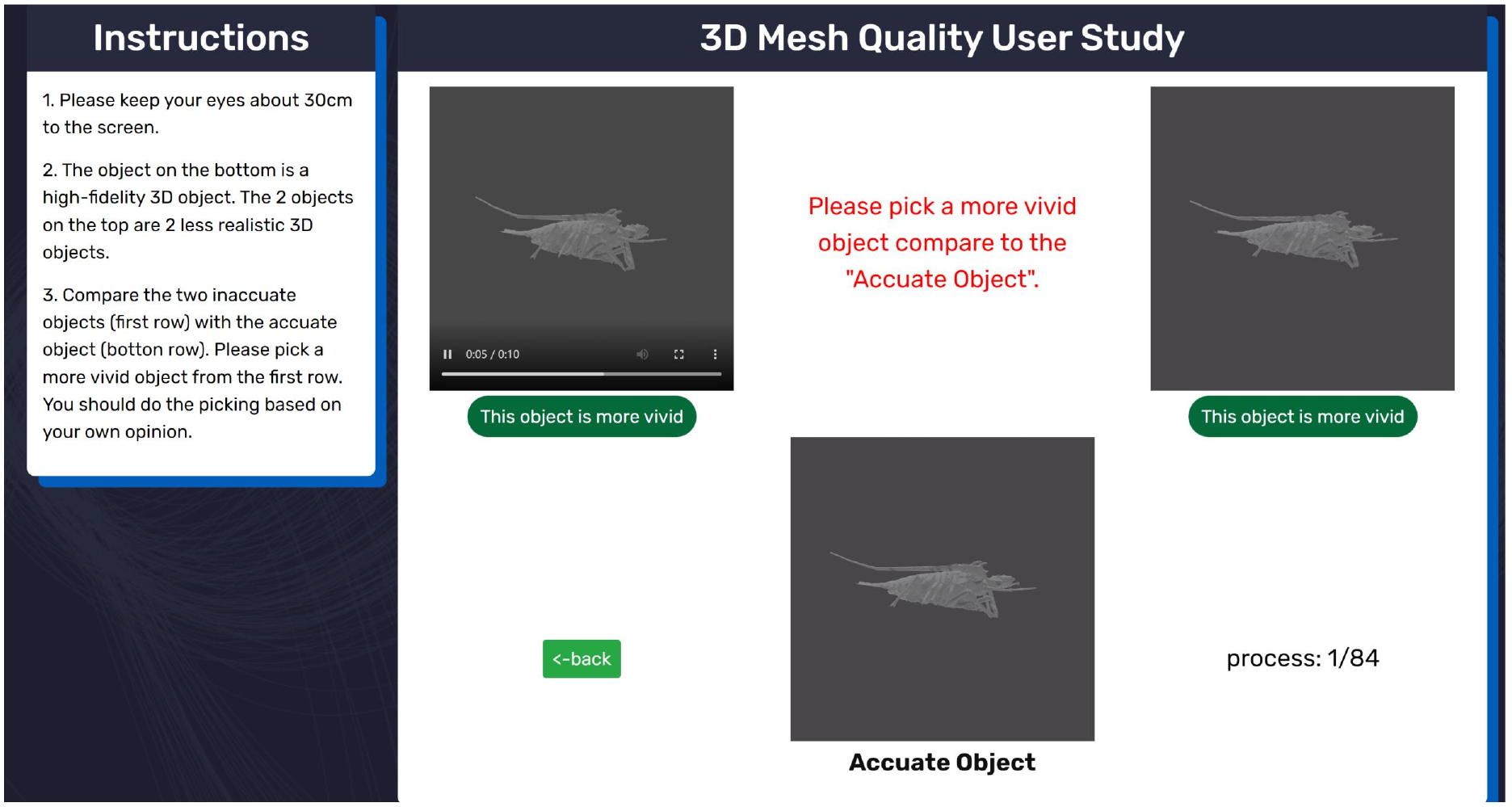}
   \caption{The panel of our online user study system. The instructions on the left contain simple instructions for the subjects. On the right side of the page, the top two videos are rendered from distorted meshes. The lower video is rendered from ground truth mesh.}
   \label{fig:panel}
\end{figure*}

\section{A Counterexample of the Original Cotan Formula not Being Positive Semidefinite}

In this section, we provide a simple mesh example to show that the original Cotan formula in \meq{2} does not guarantee to be positive semidefinite. As shown in \cref{fig:example}\textcolor{red}{a}, we reconstruct a 4-vertex mesh that is not Delauney triangulated and the mixed Voronoi areas of the vertices are not all equal. We make the two faces on the bottom ($v_1v_2v_0$ and $v_3v_0v_2$) be two congruent obtuse isosceles triangles (shown in \cref{fig:example}\textcolor{red}{b}). The apex angles of the two isosceles triangles are $\frac{2\pi}{3}$, and the base angles are $\frac{\pi}{6}$. If we make the bottom two obtuse triangles form different angles to each other, the top two triangle faces ($v_0v_1v_3$ and $v_2v_3v_1$) are always congruent isosceles triangles (as in \cref{fig:example}\textcolor{red}{c}), and their apex angles vary continuously in the range of $(0, \frac{\pi}{3})$. Here, we make the bottom two obtuse triangles form a certain angle to each other so that the apex angles of the top two triangles are equal to $\frac{\pi}{6}$, which means their base angles are $\frac{5\pi}{12}$. For simplicity, we set the equal sides of the isosceles triangles to be 1 (shown in \cref{fig:example}\textcolor{red}{a}).

Now, we calculate the DLBO metric of this reconstructed mesh using the Cotan formula in \meq{2}. First, we calculate the mixed Voronoi area for each vertex. Because of the shape symmetry, we only need to calculate the mixed Voronoi areas for vertex $v_0$ and $v_3$. The mixed Voronoi areas for vertex $v_2$ and $v_1$ are equal to $v_0$ and $v_3$, respectively. For vertex $v_0$, its mixed Voronoi area $A_0$ can be calculated as the sum of 2 times of \textcolor{Goldenrod}{yellow} area in \cref{fig:example}\textcolor{red}{b} and 1 time of \textcolor{Goldenrod}{yellow} area in \cref{fig:example}\textcolor{red}{c}, which means
\begin{equation}
\small
\begin{split}
    A_0 &= 2 \times (\frac{1}{4} \times \frac{1}{2}\cos\frac{\pi}{3}) + 1 \times (0.5 \tan\frac{\pi}{12} \times 0.5) \\
    &= \frac{4-\sqrt{3}}{8},
\end{split}
    \label{eq:mvav0}
\end{equation}
where $\frac{1}{2}\cos\frac{\pi}{3}$ is the area of the outer triangle in \cref{fig:example}\textcolor{red}{b} and $0.5 \tan\frac{\pi}{12} \times 0.5$ is the area of the \textcolor{Goldenrod}{yellow} part in \cref{fig:example}\textcolor{red}{c}. For vertex $v_3$, its mixed Voronoi area $A_3$ can be calculated as the sum of 1 time of \textcolor{ForestGreen}{green} area in \cref{fig:example}\textcolor{red}{b} and 2 times of \textcolor{ForestGreen}{green} area in \cref{fig:example}\textcolor{red}{c}, which means
\begin{equation}
\small
\begin{split}
    A_3 &= 1 \times (\frac{1}{2} \times \frac{1}{2}\cos\frac{\pi}{3}) \\
    &+ 2 \times (\frac{1}{2} \times (\sin\frac{\pi}{12}\cos\frac{\pi}{12} - 0.5 \tan\frac{\pi}{12} \times 0.5)) \\
    &= \frac{3\sqrt{3}-2}{8},
\end{split}
\label{eq:mvav3}
\end{equation}
where $\sin\frac{\pi}{12}\cos\frac{\pi}{12}$ is the area of the outer triangle in \cref{fig:example}\textcolor{red}{c}.

Second, we calculate the DLBO matrix according to \meq{2}. The DLBO matrix of the constructed mesh can be represented as
\begin{equation}
L = 
\left(
\begin{array}{llll}
\frac{w_1}{2A_0} & \frac{w_0}{2A_0} & \frac{w_3}{2A_0} & \frac{w_0}{2A_0} \\
\frac{w_0}{2A_3} & \frac{w_2}{2A_3} & \frac{w_0}{2A_3} & \frac{w_4}{2A_3} \\
\frac{w_3}{2A_0} & \frac{w_0}{2A_0} & \frac{w_1}{2A_0} & \frac{w_0}{2A_0} \\
\frac{w_0}{2A_3} & \frac{w_4}{2A_3} & \frac{w_0}{2A_3} & \frac{w_2}{2A_3}

\end{array}
\right), 
\label{eq:cot1}
\end{equation}
where 
\begin{equation}
\begin{split}
& w_0 = -(\cot\frac{5\pi}{12}+\cot\frac{\pi}{6}) = -2, \\
& w_1 = 2(\cot\frac{5\pi}{12}+\cot\frac{\pi}{6}+\cot\frac{2\pi}{3}) = 4-\frac{2\sqrt{3}}{3}, \\
& w_2 = 2(\cot\frac{5\pi}{12}+\cot\frac{\pi}{6}+\cot\frac{\pi}{6}) = 4+2\sqrt{3}, \\
& w_3 = -2\cot\frac{2\pi}{3} = \frac{2\sqrt{3}}{3}, \\
& w_4 = -2\cot\frac{\pi}{6} = -2\sqrt{3}.
\end{split}
\label{eq:cot2}
\end{equation}
Then, we can calculate the symmetric part of $L$ as 
\begin{equation}
L_{sym}=\frac{L+L^{\top}}{2}.
\label{eq:cot3}
\end{equation}
We use Wolfram Mathematica \cite{wolfram1991mathematica} to calculate the eigenvalues of $L_{sym}$. The 4 eigenvalues are
\begin{equation}
\begin{split}
& \lambda_0 = \frac{2-\frac{2\sqrt{3}}{3}}{A_0}, \\
& \lambda_1 = \frac{2+2\sqrt{3}}{A_3}, \\
& \lambda_2 = \frac{A_0+A_3-\sqrt{2(A_0^2+A_3^2)}}{A_0A_3}, \\
& \lambda_3 = \frac{A_0+A_3+\sqrt{2(A_0^2+A_3^2)}}{A_0A_3}.
\end{split}
\label{eq:ev}
\end{equation}
It is obvious that when $A_0$ and $A_3$ are both greater than 0, $\lambda_0$, $\lambda_1$, and $\lambda_3$ will be greater than 0. However, for $\lambda_2$, we have
\begin{equation}
\begin{split}
\lambda_2 &= \frac{A_0+A_3-\sqrt{2(A_0^2+A_3^2)}}{A_0A_3} \\
&= \frac{\sqrt{A_0^2+A_3^2+2A_0A_3}-\sqrt{2(A_0^2+A_3^2)}}{A_0A_3} \\
&\leq \frac{\sqrt{A_0^2+A_3^2+(A_0^2+A_3^2)}-\sqrt{2(A_0^2+A_3^2)}}{A_0A_3} \\
& = 0.
\end{split}
\label{eq:lmbd21}
\end{equation}
The equation holds if and only if $A_0 = A_3$. We know from \cref{eq:mvav0} and \cref{eq:mvav3} that $A_0 \neq A_3$. Thus, we have 
\begin{equation}
\lambda_2 < 0,
\label{eq:lmbd22}
\end{equation}
which means in the given mesh example, the original Cotan formula is not positive semidefinite.

\section{Objects and Distortions in \emph{Shape Grading}}
\cref{fig:object} shows the objects in our proposed dataset \emph{Shape Grading} and what the object numbers correspond to in \mtab{2}. We also show the distortion types that we used in our dataset and how we generate them in \cref{tab:distortion}. \cref{fig:distortion} shows examples of distorted meshes of different distortion levels in our dataset.

\begin{figure*}[t]
  \centering
    \begin{subfigure}[b]{1\textwidth}
     \centering
     \includegraphics[width=1\textwidth]{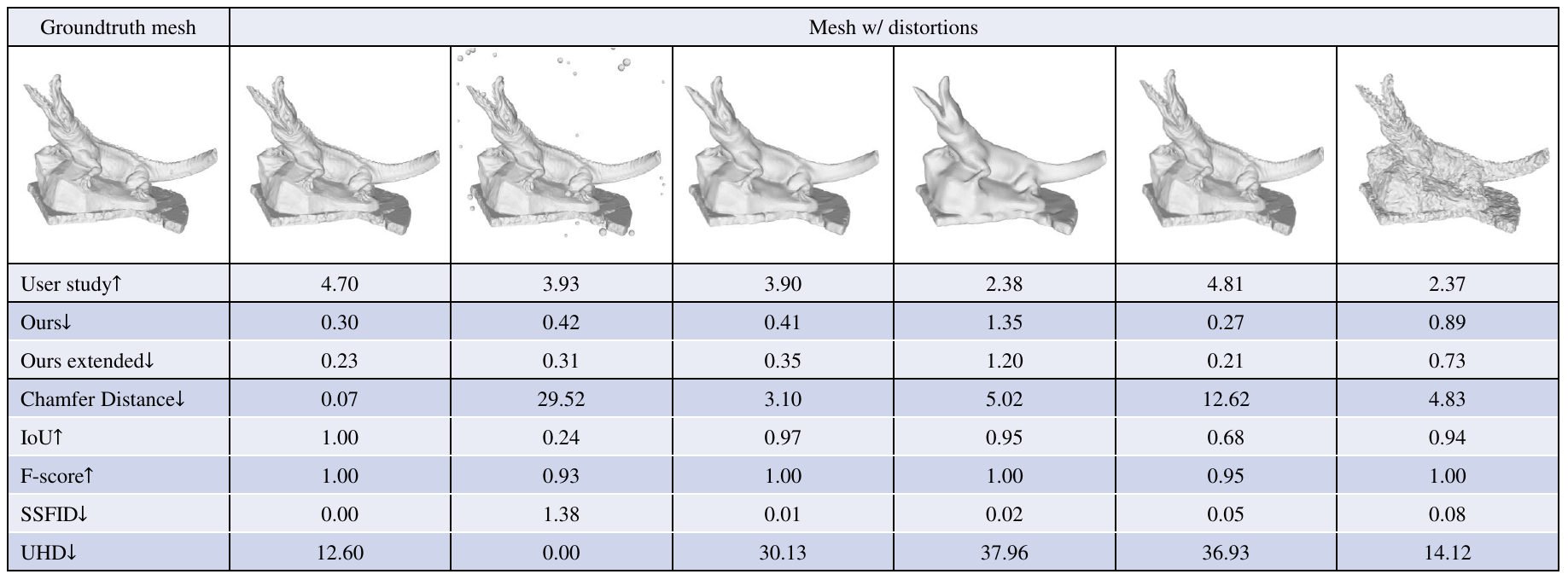}
    \end{subfigure}
    \begin{subfigure}[b]{1\textwidth}
     \centering
     \includegraphics[width=1\textwidth]{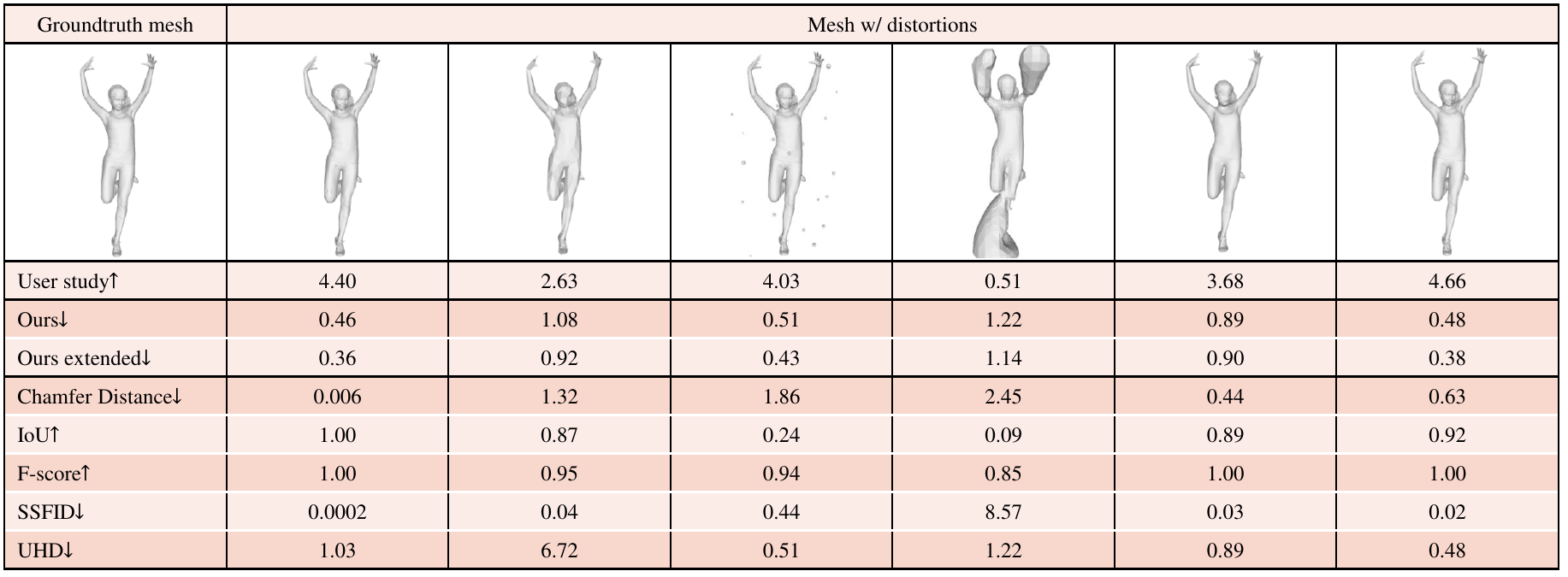}
    \end{subfigure}
    \begin{subfigure}[b]{1\textwidth}
     \centering
     \includegraphics[width=1\textwidth]{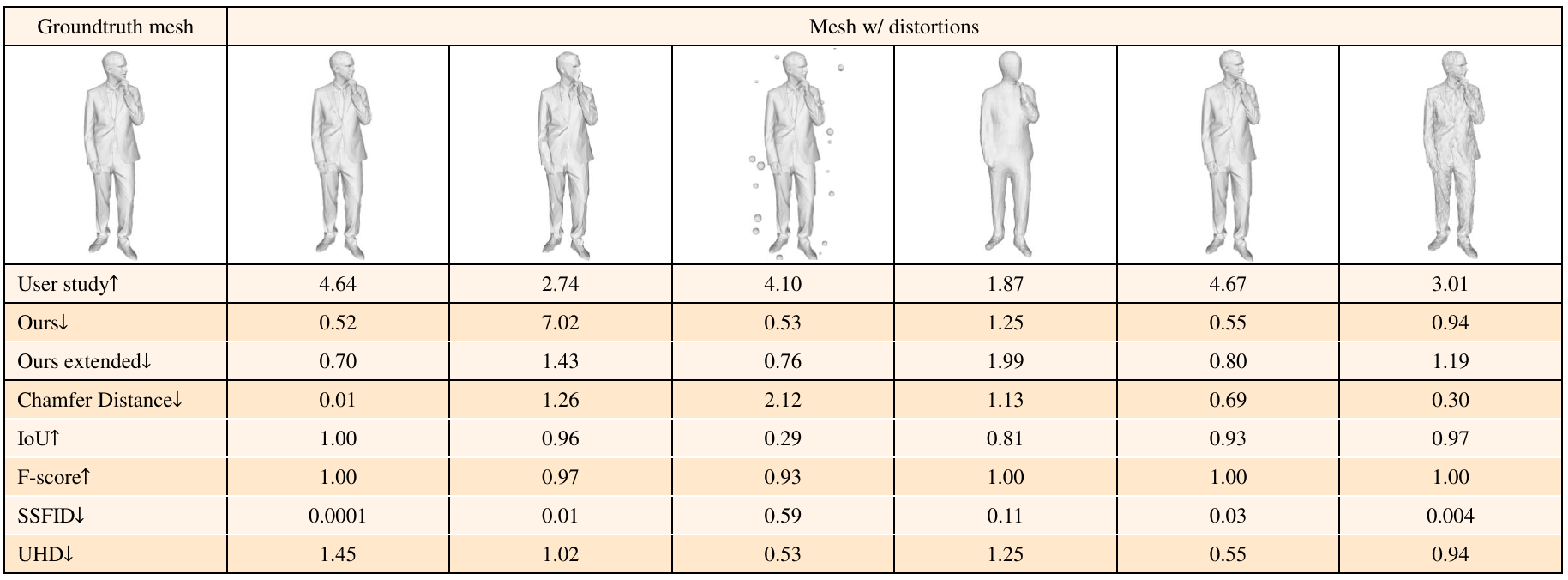}
    \end{subfigure}

  \caption{Examples in our dataset and their evaluation results using different metrics. $\downarrow$ means lower is better. $\uparrow$ means higher is better. For each object, the mesh on the top-left is the ground truth mesh, and the rest meshes are distorted meshes. The table below the meshes contains the scores they get from different metrics or from our user study. As shown in the figure, our metric aligns better with user study scores and human perception.}
  \label{fig:vis}
\end{figure*}

\begin{figure*}[t]
  \centering
   \includegraphics[width=1\linewidth]{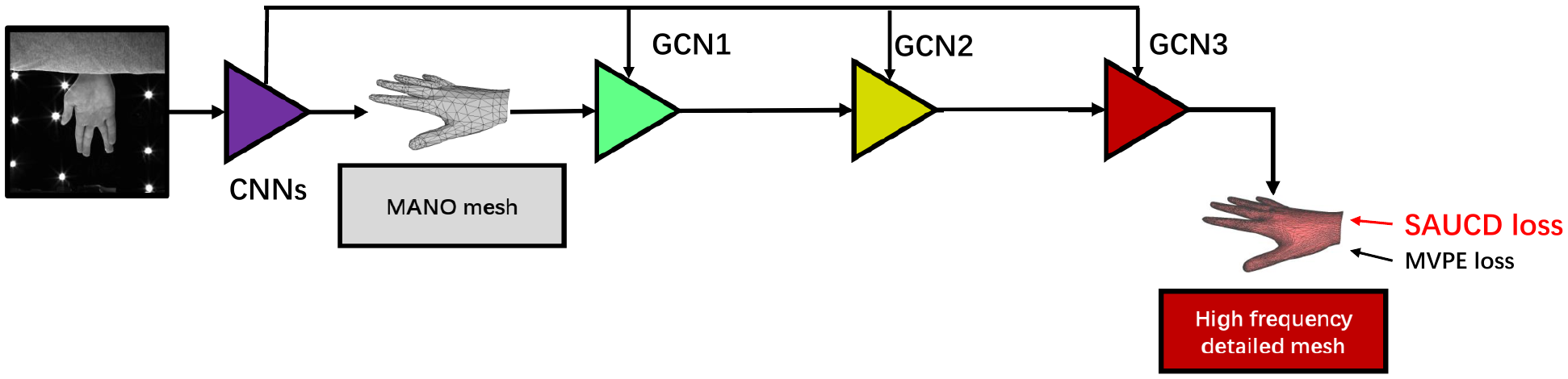}
   \caption{Network architecture used when adapting \model{} to training loss.}
   \label{fig:to_loss}
\end{figure*}

\begin{figure*}[t]
  \centering
   \includegraphics[width=1\linewidth]{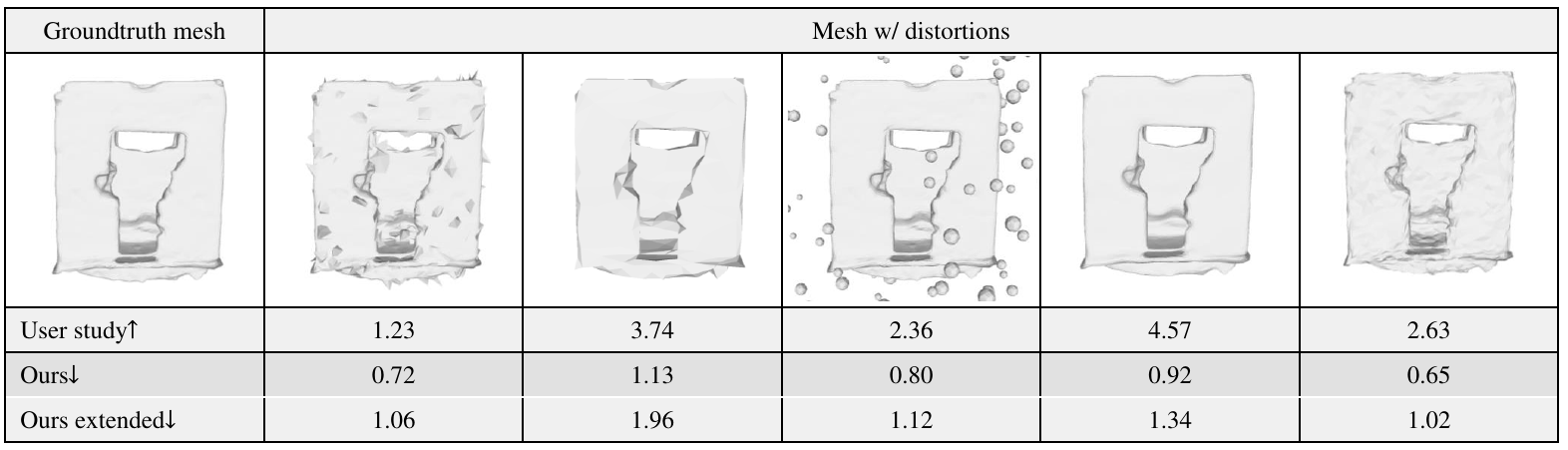}
   \caption{Failure cases. We show a case in which our metric does not provide accurate evaluations aligned with the human evaluation.}
   \label{fig:failure}
\end{figure*}

\section{Swiss System Tournament for Human Scoring}

We do a Swiss system tournament for human scoring in \msec{4.1}. The tournament has 6 rounds. To begin with, all 28 meshes are set to 0 points. In the first round, the 28 meshes are randomly sorted and we form the adjacent meshes into pairs (the 1st and 2nd meshes form a pair, the 3rd and 4th meshes form another pair, \etc{}). Together, we have 14 pairs. For each pair, we ask the subject which one is closer to the ground truth. The mesh that the subject picked will be added 1 point. From the 2nd to the 6th round, for each round, we first sort the meshes by their current score from low to high, and we also make pairs with adjacent meshes in the sorted mesh array, like what we did in the first round. The mesh closer to ground truth will be added 1 point. The scores of the meshes after 6 rounds are their scores graded by this subject. \cref{fig:panel} shows the panel of our online human scoring page.

\section{More Examples and Evaluation Results}
We show more examples in our dataset and evaluation results using different metrics in \cref{fig:vis}. Compared to previous methods, our provided metrics generally align better with the human evaluation of mesh shape similarity.

\section{Implementation Details on Adapting \model{} to Training Loss}

We adapt \model{} to a topology Laplacian version. Specifically, we replace the Laplacian matrix defined in the main paper Eq.(4) to $L=D-A$ defined in \cite{chung1997spectral}, where $D$ is the degree matrix of the mesh graph, and $A$ is the adjacency matrix of the mesh graph. By making the change, we can avoid calculating a different SVD decomposition in every training iteration when mesh vertex locations change. Our network is designed as \cref{fig:to_loss}. The input image first goes through a feature extraction CNNs network to get image features, and uses that feature to generate MANO~\cite{MANO:SIGGRAPHASIA:2017} mesh. Then, we use features from CNNs network and 3 resolution levels of Graph Convolution Networks (GCN) to reconstruct the mesh details. In the main paper Fig. 8, we compare the results using only MVPE loss (w/o \model{} loss column) and using both MVPE and \model{} loss (w/ \model{} loss column). In this experiment, we use EfficientNet~\cite{tan2019efficientnet} and GCN similar to ~\cite{kolotouros2019convolutional}.

\section{Failure Cases}
We also show a case that our metric does not provide accurate evaluations aligned with the human evaluation in \cref{fig:failure}.

\section{Discussions of Future Works}
In future work, we plan to dig deeper into understanding human sensitivity to frequency changes. To enhance the robustness and applicability of our approach, we plan to expand our dataset to include a wider range of distortions and objects. While our current methods are effective on general 3D meshes, we recognize the importance of developing specialized versions for particular areas of 3D reconstruction, such as human body~\cite{gong2022self,gong2023progressive,luan2021pc}, human face~\cite{gao2020semi,gao2023learning}, human hand~\cite{luan2023high}, or 3D volumetric approaches~\cite{pan2023cycle, pan2023data}. Furthermore, the frequency method holds promise for extension into 2D domains, including the analysis and generation of images ~\cite{xie2023shisrcnet, zhai2023towards, lou2023min, zhang2021task, yu2022reducing, yu2023deep}, as well as videos~\cite{zhai2023language, zhu2021enriching, chen2022learning, zhu2022learning, zhai2020two}. These future works will not only refine our understanding of human perception alignment but also broaden the potential applications of our research in various fields.

{
    \small
    \bibliographystyle{ieeenat_fullname}
    \bibliography{saucd}
}


\end{document}

%% file: preamble.tex
\usepackage[dvipsnames]{xcolor}


%% file: tables-metrics.tex
\begin{table*}[t]
	
    \hfil
    \begin{minipage}{0.82\linewidth}
    \centering
    \subfloat{
    \resizebox{\linewidth}{!}{%
    \begin{tabular}{|l|c|c|c|c|c|c|c|c|c|c|c|c|>{\columncolor{gray!20}}c|}
    \hline 
    \rowcolor{red!15}\diagbox{Metrics}{Object No.} & 1 & 2 & 3 & 4 & 5 & 6 & 7 & 8 & 9 & 10 & 11 & 12 & Overall\tabularnewline
    \hline\hline
    Chamfer Distance~\cite{borgefors1984CD} & 0.54 & 0.15 & -0.10 & 0.57 & -0.06 & -0.12 & -0.20 & 0.07 & 0.04 & 0.30 & -0.20 & 0.17 & 0.097 \tabularnewline
    Point-to-Surface & 0.45 & 0.19 & -0.04 & \underline{0.66} & -0.08 & -0.25 & -0.32 & -0.20 & 0.01 & 0.13 & -0.21 & -0.12 & 0.017 \\
    Normal Difference & 0.46 & 0.11 & 0.06 & 0.28 & 0.11 & 0.21 & 0.29 & 0.47 & 0.27 & 0.39 & 0.11 & 0.27 & 0.253 \\
    IoU~\cite{henderson2018iou} & 0.60 & \underline{0.63} & 0.01 & 0.51 & 0.30 & 0.02 & -0.07 & 0.20 & 0.14 & 0.47 & -0.09 & -0.01 & 0.225 \tabularnewline
    F-score~\cite{wang2018fscore} & 0.58 & 0.09 & 0.05 & 0.33 & 0.03 & 0.06 & 0.16 & 0.34 & 0.27 & 0.25 & 0.01 & \underline{0.34} & 0.208 \tabularnewline
    SSFID~\cite{wu2022ssfid} & 0.71 & \textbf{0.74} & -0.04 & \textbf{0.74} & \textbf{0.39} & 0.24 & 0.13 & 0.32 & 0.25 & 0.64 & 0.25 & -0.02 & 0.363 \tabularnewline
    UHD~\cite{wu2020uhd} & 0.29 & 0.22 & 0.11 & 0.15 & -0.04 & 0.18 & 0.41 & 0.55 & 0.13 & 0.18 & 0.25 & 0.33 & 0.231  \tabularnewline
    \hline 
    \model{} (Ours) & \underline{0.73} & 0.21 & \textbf{0.60} & 0.63 & 0.31 & \underline{0.51} & \textbf{0.83} & \underline{0.65} & \textbf{0.77} & \textbf{0.80} & \textbf{0.69} & 0.08 & \underline{0.567} \tabularnewline
    Adjusted \model{} (Ours) & \textbf{0.79} & 0.19 & \underline{0.56} & 0.64 & \underline{0.36} & \textbf{0.54} & \underline{0.79} & \textbf{0.76} & \underline{0.75} & \underline{0.77} & \underline{0.67} & \textbf{0.36} & \textbf{0.598} \tabularnewline
    \hline 
    \end{tabular}%
    }
    }
    
    {\small\textcolor{red}{a. Pearson's linear correlation coefficient.}}
	\end{minipage}
	\hfil
 
	\vspace{2 pt}
 
	\hfil
	\begin{minipage}{0.82\linewidth}
		\centering
		\subfloat{
	\resizebox{\linewidth}{!}{%
    \begin{tabular}{|l|c|c|c|c|c|c|c|c|c|c|c|c|>{\columncolor{gray!20}}c|}
    \hline 
    \rowcolor{blue!15}\diagbox{Metrics}{Object No.} & 1 & 2 & 3 & 4 & 5 & 6 & 7 & 8 & 9 & 10 & 11 & 12 & Overall\tabularnewline
    \hline\hline
    Chamfer Distance~\cite{borgefors1984CD} & 0.33 & 0.14 & -0.09 & 0.43 & -0.08 & -0.06 & -0.15 & 0.17 & -0.04 & 0.24 & -0.16 & 0.22 & 0.079 \tabularnewline
    Point-to-Surface & 0.42 & 0.39 & 0.14 & 0.59 & 0.11 & 0.05 & -0.10 & 0.20 & 0.18 & 0.40 & -0.11 & 0.18 & 0.205 \\
    Normal Difference & 0.44 & 0.22 & 0.33 & 0.42 & 0.19 & 0.29 & 0.33 & 0.56 & 0.33 & 0.32 & 0.21 & 0.34 & 0.331 \\
    IoU~\cite{henderson2018iou} & 0.57 & \underline{0.61} & 0.28 & 0.50 & 0.36 & 0.21 & 0.12 & 0.31 & 0.262 & 0.56 & 0.03 & 0.30 & 0.342 \tabularnewline
    F-score~\cite{wang2018fscore} & 0.47 & 0.25 & 0.20 & 0.52 & 0.21 & 0.11 & 0.07 & 0.36 & 0.30 & 0.42 & -0.01 & 0.35 & 0.27 \tabularnewline
    SSFID~\cite{wu2022ssfid} & 0.63 & \textbf{0.81} & 0.28 & \textbf{0.70} & 0.33 & 0.23 & 0.10 & 0.33 & 0.32 & 0.65 & 0.16 & 0.34 & 0.407  \tabularnewline
    UHD~\cite{wu2020uhd} & 0.38 & 0.20 & 0.11 & 0.32 & 0.13 & 0.35 & 0.41 & 0.60 & 0.06 & 0.27 & 0.37 & \underline{0.35} & 0.296 \tabularnewline
    \hline 
    \model{} (Ours) & \underline{0.79} & 0.25 & \textbf{0.57} & 0.59 & \underline{0.36} & \underline{0.56} & \textbf{0.83} & \underline{0.79} & \underline{0.69} & \textbf{0.69} & \textbf{0.83} & 0.24 & \underline{0.598} \tabularnewline
    Adjusted \model{} (Ours) & \textbf{0.83} & 0.21 & \underline{0.55} & \underline{0.59} & \textbf{0.38} & \textbf{0.60} & \underline{0.82} & \textbf{0.80} & \textbf{0.69} & \underline{0.68} & \underline{0.75} & \textbf{0.42} & \textbf{0.611} \tabularnewline
    \hline 
    \end{tabular}%
    }
		}
  
  {\small\textcolor{blue}{b. Spearman's rank order correlation coefficient.}}
	\end{minipage}
	\hfil
	 
	\vspace{2 pt}
 
	\hfil
	\begin{minipage}{0.82\linewidth}
		\centering
		\subfloat{
	\resizebox{\linewidth}{!}{%
    \begin{tabular}{|l|c|c|c|c|c|c|c|c|c|c|c|c|>{\columncolor{gray!20}}c|}
    \hline 
    \rowcolor{cyan!15}\diagbox{Metrics}{Object No.} & 1 & 2 & 3 & 4 & 5 & 6 & 7 & 8 & 9 & 10 & 11 & 12 & Overall\tabularnewline
    \hline\hline
    Chamfer Distance~\cite{borgefors1984CD} & 0.25 & 0.14 & -0.08 & 0.31 & -0.04 & -0.02 & -0.09 & 0.15 & 0.013 & 0.19 & -0.07 & 0.22 & 0.080  \tabularnewline
    Point-to-Surface & 0.33 & 0.30 & 0.07 & \underline{0.45} & 0.10 & 0.08 & -0.03 & 0.17 & 0.13 & 0.30 & -0.01 & 0.16 & 0.171 \\
    Normal Difference & 0.34 & 0.16 & 0.17 & 0.31 & 0.18 & 0.22 & 0.26 & 0.44 & 0.25 & 0.23 & 0.16 & 0.27 & 0.250 \\
    IoU~\cite{henderson2018iou} & 0.42 & \underline{0.44} & 0.24 & 0.37 & \underline{0.28} & 0.22 & 0.14 & 0.26 & 0.20 & 0.41 & 0.10 & 0.23 & 0.275 \tabularnewline
    F-score~\cite{wang2018fscore} & 0.37 & 0.17 & 0.14 & 0.42 & 0.15 & 0.11 & 0.09 & 0.28 & 0.23 & 0.34 & 0.01 & \textbf{0.30} & 0.216\tabularnewline
    SSFID~\cite{wu2022ssfid} & 0.48 & \textbf{0.62} & 0.24 & \textbf{0.51} & 0.25 & 0.24 & 0.12 & 0.29 & 0.26 & \textbf{0.48} & 0.17 & 0.23 & 0.322 \tabularnewline
    UHD~\cite{wu2020uhd} & 0.27 & 0.13 & 0.07 & 0.22 & 0.09 & 0.26 & 0.29 & 0.42 & 0.048 & 0.19 & 0.28 & 0.24 & 0.209 \tabularnewline
    \hline 
    \model{} (Ours) & \underline{0.60} & 0.16 & \textbf{0.42} & 0.41 & 0.27 & \underline{0.45} & \textbf{0.65} & \underline{0.57} & \textbf{0.55} & \underline{0.47} & \textbf{0.60} & 0.19 & \underline{0.445}  \tabularnewline
    Adjusted \model{} (Ours) & \textbf{0.64} & 0.14 & \underline{0.40} & 0.41 & \textbf{0.29} & \textbf{0.48} & \underline{0.63} & \textbf{0.59} & \underline{0.55} & 0.45 & \underline{0.57} & \underline{0.29} & \textbf{0.453} \tabularnewline
    \hline 
    \end{tabular}%
    }
		}
  
  {\small\textcolor{cyan}{c. Kendall's rank order correlation coefficient.}}
	\end{minipage}
	\hfil

        \vspace{2pt}
        
	\caption{Correlations between different metrics and human annotation. ``\model{}" is our basic version metric. ``Adjusted \model{}" is the human-adjusted version of our metric. The ranges of all three correlation coefficients are {$[-1, 1]$}, and the higher the better. }
	\label{tab:correlation}
\end{table*}

%% file: tables-ablation.tex
\begin{table*}[t]
\begin{minipage}[h]{0.3\linewidth}
\centering
\resizebox{\linewidth}{!}{%
    \begin{tabular}{|l|>{\columncolor{red!15}}c|>{\columncolor{blue!15}}c|>{\columncolor{cyan!15}}c|}
    \hline
    Frequency band & PLCC $\uparrow$ & SROCC $\uparrow$ & KROCC $\uparrow$ \tabularnewline
    \hline\hline
    $[0, 0.001)$ & 0.434 & 0.515 & 0.376 \\
    $[0.001, 0.003)$ & 0.240 & 0.409 & 0.281 \\
    $[0.003, 0.01)$ & 0.255 & 0.455 & 0.340 \\
    $[0.01, 0.03)$ & 0.421 & 0.528 & 0.391 \\
    $[0.03, 0.1)$ & 0.287 & 0.351 & 0.250 \\
    $[0.1, \infty)$ & 0.318 & 0.192 & 0.155 \\
    \hline
    $[0, \infty)$ & \textbf{0.567} & \textbf{0.598} & \textbf{0.445} \\
    \hline
    \end{tabular}%
}
\caption{Results when building metrics using each frequency band separately. The bottom row is our proposed metric.}
\label{tab:freqband}
\end{minipage}
\hfill
\begin{minipage}[h]{0.31\linewidth}
\centering
\resizebox{\linewidth}{!}{%
    \begin{tabular}{|l|>{\columncolor{red!15}}c|>{\columncolor{blue!15}}c|>{\columncolor{cyan!15}}c|}
    \hline
    Pruning Portion & PLCC $\uparrow$ & SROCC $\uparrow$ & KROCC $\uparrow$ \tabularnewline
    \hline\hline
    $0\%$ & 0.513 & 0.549 & 0.393 \\
    $0.1\%$ & \textbf{0.567} & 0.598 & \textbf{0.445} \\
    $1\%$ & 0.554 & \textbf{0.602} & 0.462 \\
    $10\%$ & 0.517 & 0.581 & 0.442 \\
    $20\%$ & 0.503 & 0.587 & 0.445 \\
    \hline
    \end{tabular}%
}
\caption{Results with different pruning portions. The metric achieves better results with pruning portion to be $0.1\%$ or $1\%$. We use pruning portion as $0.1\%$ in our design.}
\label{tab:prune}
\end{minipage}
\hfill
\begin{minipage}[h]{0.36\linewidth}
\centering
\resizebox{\linewidth}{!}{%
    \begin{tabular}{|l|>{\columncolor{red!15}}c|>{\columncolor{blue!15}}c|>{\columncolor{cyan!15}}c|}
    \hline
    Modules & PLCC $\uparrow$ & SROCC $\uparrow$ & KROCC $\uparrow$ \tabularnewline
    \hline\hline
    Topology Laplacian~\cite{chung1997spectral}
     & 0.298 & 0.327 & 0.235 \\
    Cotan formula~\cite{meyer2003dlbo}
     & 0.417 & 0.470 & 0.340 \\
    \hline
    Energy difference 
     & 0.268 & 0.315 & 0.215 \\
    \hline
    w/o normalization
     & 0.257 & 0.507 & 0.353 \\
    Spatial normalization
     & 0.269 & 0.542 & 0.392 \\
    
    \hline
    Ours & \textbf{0.567} & \textbf{0.598} & \textbf{0.445} \\
    \hline
    \end{tabular}%
}
\caption{Module replacement. We replace each module of our metric with alternative designs to verify the design of each module.}
\label{tab:module}
\end{minipage}
\end{table*}